\documentclass[10pt,twocolumn,letterpaper]{article}

\usepackage{cvpr}              % To produce the CAMERA-READY version

\usepackage{mathtools}
\usepackage{colortbl}
\usepackage{graphicx}
\usepackage{subcaption}

\usepackage{caption}

\usepackage{amsfonts}       % blackboard math symbols
\usepackage{amsmath}

\usepackage{url}            % simple URL typesetting
\usepackage{booktabs}
\usepackage{algorithm}
\usepackage{algorithmic}
\usepackage{multirow}
\usepackage{tabularx}
\usepackage{amsfonts,bm}
\usepackage{threeparttable}
\usepackage{upgreek}
\usepackage{pifont}%
\usepackage{bbding}
\usepackage[normalem]{ulem}
\usepackage{makecell}
\usepackage{color}
\usepackage{xcolor}
\usepackage{colortbl}
\usepackage{wrapfig}
\usepackage{soul}
\usepackage{wasysym}
\usepackage{xspace}

\usepackage{enumitem}
\usepackage[accsupp]{axessibility}
\usepackage{comment}

\definecolor{Gray}{gray}{0.86}

\definecolor{cvprblue}{rgb}{0.21,0.49,0.74}
\definecolor{Gray}{gray}{0.86}

\newcommand{\rblue}{\rowcolor{blue!10}}

\usepackage[pagebackref,breaklinks,colorlinks,citecolor=cvprblue]{hyperref}

\usepackage{pifont}
\def\paperID{5252} % *** Enter the Paper ID here
\def\confName{CVPR}
\def\confYear{2025}

\title{Building Vision Models upon Heat Conduction}

\author{
Zhaozhi Wang\textsuperscript{$1,2$}\thanks{Equal contribution.}, Yue Liu\textsuperscript{$1$}\footnotemark[1], Yunjie Tian\textsuperscript{$1$}, Yunfan Liu\textsuperscript{$1$}, Yaowei Wang\textsuperscript{$2,3$}, Qixiang Ye\textsuperscript{$1,2$}\thanks{Corresponding author.}
\\
\\
\textsuperscript{$1$}University of Chinese Academy of Sciences \quad \textsuperscript{$2$}Peng Cheng Lab \\
\textsuperscript{$3$}Harbin Institute of Technology (Shenzhen) \\
{\tt\small \{wangzhaozhi22,liuyue171,tianyunjie19\}@mails.ucas.ac.cn}\\\quad{\tt\small yunfan.liu@ucas.ac.cn}\quad{\tt\small wangyw@pcl.ac.cn}\quad{\tt\small qxye@ucas.ac.cn}
}

\begin{document}
\maketitle
\begin{abstract}
%
%A fundamental problem in learning robust and expressive visual representations lies in efficiently estimating the spatial relationships of visual semantics throughout the entire image. In this study, we propose vHeat, a novel vision backbone model that simultaneously achieves both high computational efficiency and global receptive field. The essential idea, inspired by the physical principle of heat conduction, is to conceptualize image patches as heat sources and model the calculation of their correlations as the diffusion of thermal energy. This mechanism is incorporated into deep models through the newly proposed module, the Heat Conduction Operator (HCO), which is physically plausible and can be efficiently implemented using DCT and IDCT operations with a complexity of $\mathcal{O}(N^{1.5})$. Extensive experiments demonstrate that vHeat surpasses Vision Transformers (ViTs) across various vision tasks, while also providing higher inference speeds, reduced FLOPs, and lower GPU memory usage for high-resolution images.
%
%Efficiently estimating the spatial relationships of visual semantics throughout an image is a fundamental challenge when learning robust and expressive visual representations.
%
Visual representation models leveraging attention mechanisms are challenged by significant computational overhead, particularly when pursuing large receptive fields. In this study, we aim to mitigate this challenge by introducing the Heat Conduction Operator (HCO) built upon the physical heat conduction principle. HCO conceptualizes image patches as heat sources and models their correlations through adaptive thermal energy diffusion, enabling robust visual representations. HCO enjoys a computational complexity of $O(N^{1.5})$, as it can be implemented using discrete cosine transformation (DCT) operations. HCO is plug-and-play, combining with deep learning backbones produces visual representation models (termed vHeat) with global receptive fields. Experiments across vision tasks demonstrate that, beyond the stronger performance, vHeat achieves up to a 3$\times$ throughput, 80\% less GPU memory allocation, and 35\% fewer computational FLOPs compared to the Swin-Transformer. Code is available at \href{https://github.com/MzeroMiko/vHeat}{\color{magenta}https://github.com/MzeroMiko/vHeat} and \href{https://openi.pcl.ac.cn/georgew/vHeat}{\color{magenta}{https://openi.pcl.ac.cn/georgew/vHeat}}. 

\end{abstract}    
\section{Introduction}

Convolutional Neural Networks (CNNs)~\citep{AlexNet2012, Resnet2016} have been the cornerstone of visual representation since the advent of deep learning, exhibiting remarkable performance across vision tasks. 
However, the reliance on local receptive fields and fixed convolutional operators imposes constraints, particularly in capturing long-range and complex dependencies within images~\citep{luo2016understanding}.
These limitations have motivated significant interest in developing alternative visual representation models, including architectures based on ViTs~\citep{ViT2021,Swin2021} and State Space Models~\citep{zhu2024vision,liu2024vmamba}. 
Despite their effectiveness, these models continue to face challenges, including relatively high computational complexity and a lack of interpretability.

\begin{figure*}
    \begin{center}
        \includegraphics[width=0.8\textwidth]{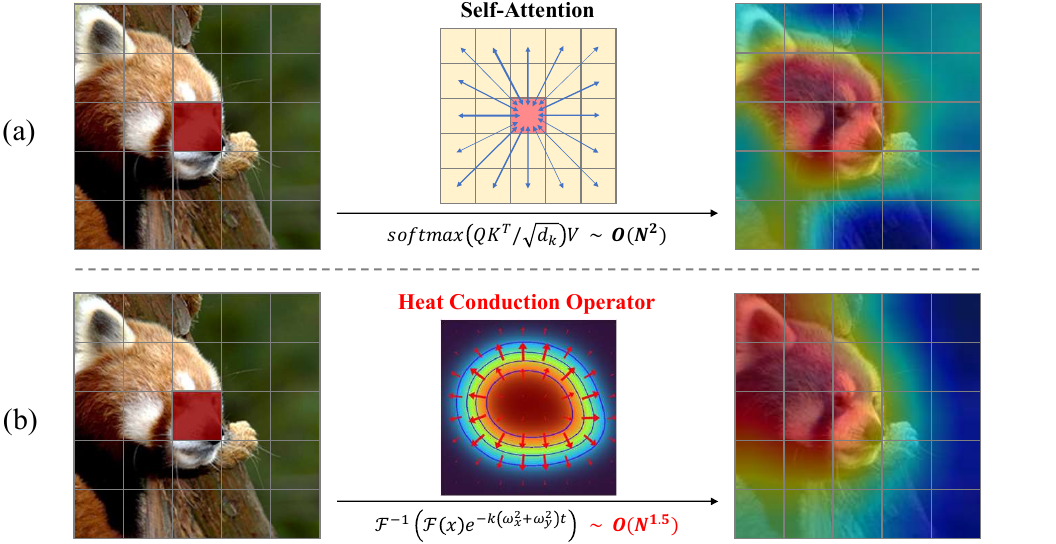}
        \caption{Comparison of information conduction mechanisms: self-attention \textit{vs.} heat conduction. (a) The self-attention operator uniformly ``conducts'' information from a pixel to all other pixels, resulting in $\mathcal{O}(N^2)$ complexity. (b) The heat conduction operator (HCO) conceptualizes the center pixel as the heat source and conducts information propagation through DCT ($\mathcal{F}$) and IDCT ($\mathcal{F}^{-1}$), which enjoys interpretability, global receptive fields, and $\mathcal{O}(N^{1.5})$ complexity.}
        \label{fig:conduction_comp}
    \end{center}
\end{figure*}

When addressing these limitations, we draw inspiration from the field of heat conduction~\citep{widder1976heat}, where \textit{spatial locality} is crucial for the transfer of thermal energy due to the collision of neighboring particles.
Notably, analogies can be drawn between the principles of heat conduction and the propagation of visual semantics within the spatial domain, as adjacent image regions in a certain scale tend to contain related information or share similar characteristics.
Leveraging these connections, we introduce \textbf{vHeat}, a physics-inspired vision representation model that conceptualizes image patches as \textit{heat sources} and models the calculation of their correlations as the diffusion of thermal energy.

To integrate the principle of heat conduction into deep networks, we first derive the general solution of heat conduction in 2D space and extend it to multiple dimensions, corresponding to feature channels.
Based on this general solution, we design the \textbf{Heat Conduction Operator (HCO)}, which simulates the propagation of visual semantics across image patches along multiple dimensions.
%following the physical law of heat conduction. 
%
Notably, we demonstrate that HCO can be approximated through 2D (inverse) discrete cosine transformation (DCT/IDCT), effectively reducing the computational complexity to $\mathcal{O}(N^{1.5})$, Fig.~\ref{fig:conduction_comp}. 
This improvement boosts both training and testing efficiency due to the high parallelizability of DCT and IDCT operations.
Furthermore, as each element in the frequency domain obtained by DCT incorporates information from all patches in the image space, vHeat can establish long-range feature dependencies and achieve global receptive fields. 
%%%%%%%%%%%%%%%%%%%%%%%%%%%%%%%%%%%%%%%%%%%%%%%%%%%%%%%%%%%%%%%%%%%
%这里可以加两句：引入导热系数，说明导热系数的可学习保证了模型的非线性和表征能力，
%%%%%%%%%%%%%%%%%%%%%%%%%%%%%%%%%%%%%%%%%%%%%%%%%%%%%%%%%%%%%%%%%%%
To enhance the representation adaptability of vHeat, we propose learnable frequency value embeddings (FVEs) to characterize the frequency information and predict the thermal diffusivity of visual heat conduction.

We develop a family of vHeat models (\textit{i.e.}, vHeat-Tiny/Small/Base), and extensive experiments are conducted to demonstrate their effectiveness in diverse visual tasks. 
Compared to benchmark vision backbones with various architectures (\textit{e.g.}, ConvNeXt~\citep{liu2022convnet}, Swin~\citep{Swin2021}, and Vim~\citep{zhu2024vision}), vHeat consistently achieves superior performance on image classification, object detection, and semantic segmentation across model scales. 
Specifically, vHeat-Base achieves a $84.0\%$ top-1 accuracy on ImageNet-1K, surpassing Swin by $0.5\%$, with a throughput exceeding that of Swin by a substantial margin over $40\%$ ($661$ \textit{vs.} $456$). 
To explore the generalization of vHeat, we've also validated its superiority on robustness evaluation benchmarks and low-level vision tasks. 
Besides, due to the $\mathcal{O}(N^{1.5})$ complexity of HCO, vHeat enjoys considerably lower computational cost compared to ViT-based models, demonstrating significantly reduced FLOPs and GPU memory requirements, and higher throughput as image resolution increases.
In particular, when the input image resolution increases to $768\times 768$, vHeat-Base achieves a 3$\times$ throughput compared to Swin, with 80\% less GPU memory allocation and 35\% fewer computational FLOPs, as shown in Fig.~\ref{fig:cost}.
%
% For image recognition on ImageNet-1K, the family of vHeat models exhibits significant performance advantages at lower computational cost compared with ResNet~\citep{Resnet2016}, ViT~\citep{ViT2021}, and Swin~\citep{Swin2021}. 
% %
% In downstream vision tasks, vHeat-Tiny/Small/Base achieves $45.1\%/46.8\%/47.7\%$ mAP on COCO~\citep{COCO2014} with the Mask RCNN detector ($1\times$ training schedule) and $46.9\%/49.0\%/49.6\%$ mIoU (Single-Scale) on ADE20K~\citep{zhou2017scene} using UperNet with $512\times512$ input, demonstrating its great potential to serve as a powerful backbone model. 
% %
% Furthermore, compared to ViTs~\citep{Swin2021}, vHeat exhibits significantly lower FLOPs and GPU memory growth, and faster throughput when the image resolution grows.

%%%%%%%%%%%%%%%%%%%%%%%%%%%%%%%%%%%%%%%%%%%%%%%%%%%%%%%%%%%%%%%%%%%%%%%%%%%%%%%%
The contributions of this study are summarized as follows:
\begin{itemize}

    \item We propose vHeat, a vision backbone model inspired by the physical principle of heat conduction, which simultaneously achieves global receptive fields, low computational complexity, and high interpretability.
    
    \item We design the Heat Conduction Operator (HCO), a physically plausible module conceptualizing image patches as heat sources, predicting adaptive thermal diffusivity by FVEs, and transferring information following the principles of heat conduction. 
    %
    %HCO is plug-and-play, as it can be integrated into other vision backbones by replacing the attention module.
    %combined with deep networks by simply replacing existing operators.
    
    \item Without bells and whistles, vHeat achieves promising performance in vision tasks including image classification, object detection, and semantic segmentation. It also enjoys higher inference speeds, reduced FLOPs, and lower GPU memory usage for high-resolution images.
     
\end{itemize}
%%%%%%%%%%%%%%%%%%%%%%%%%%%%%%%%%%%%%%%%%%%%%%%%%%%%%%%%%%%%%%%%%%%%%%%%%%%%%%%%

\section{Related Work}\label{related_work}

%
%In the past two decades, two prevalent categories of vision models, \textit{i.e.}, CNNs~\citep{AlexNet2012, vgg, googlenet, Resnet2016, EfficientNet2019} and ViTs~\citep{ViT2021, Swin2021, pvt, dong2022cswin, dai2021coatnet, hivit}, dominate the computer vision area. 
%

\begin{figure*}[!htb]
\centering
\begin{center}
\includegraphics[width=0.85\textwidth]{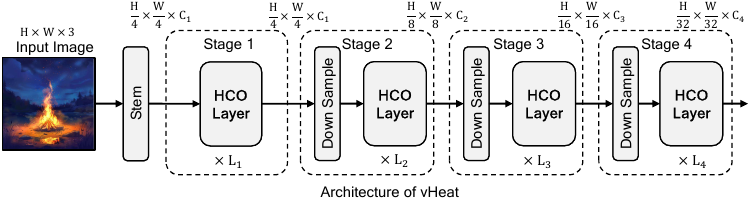}
\caption{The network architecture of vHeat. Following the traditional principles of visual model design, we built vHeat with 4 HCO blocks, connected by downsampling layers in between.}
\label{fig:vheat_arch}
\end{center}
\vspace{-15pt}
\end{figure*}

\noindent\textbf{Convolution Neural Networks.}
CNNs have been landmark models in the history of visual perception ~\citep{lecun1998gradient, AlexNet2012}.
%Early CNN-based models~\citep{lecun1998gradient, AlexNet2012} are designed for basic tasks~\citep{lecun1989backpropagation, zhang2015character}.
%
The distinctive characteristics of CNNs are encapsulated in the convolution kernels, which enjoy high computational efficiency given specifically designed GPUs. 
With the aid of powerful GPUs and large-scale datasets~\citep{ImageNet2009}, increasingly deeper~\citep{vgg, googlenet, Resnet2016, densenet} and efficient models~\citep{howard2017mobilenets, EfficientNet2019, yang2021focal, radosavovic2020designing} have been proposed for higher performance across a spectrum of vision tasks.
Numerous modifications have been made to the convolution operators to improve its capacity~\citep{chollet2017xception}, efficiency~\citep{ hua2018pointwise, dilation} and adaptability~\citep{dai2017deformable,InternImage24}. 
Nevertheless, the born limitation of local receptive fields remains. Recently developed large convolution kernels~\citep{replknet} took a step towards large receptive fields, but experienced difficulty in handling high-resolution images.  

\noindent\textbf{Vision Transformers.} 
Built upon the self-attention operator~\citep{Attention2017}, ViTs have the born advantage of building global feature dependency. Based on the learning capacity of self-attention across all image patches, ViTs has been the most powerful vision model ever, given a large dataset for pre-training~\citep{ViT2021,DeiT2021,beitv2}. The introduction of hierarchical architectures~\citep{Swin2021, dong2022cswin, pvt, container, hivit, tian2023integrally, dai2021coatnet, ding2022davit, zhao2022graformer} further improves the performance of ViTs. 
The Achilles' Heel of ViTs is the $\mathcal{O}(N^2)$ computational complexity, which implies substantial computational overhead given high-resolution images. Great efforts have been made to improve model efficiency by introducing window attention, linear attention and cross-covariance attention operators~\citep{wang2020linformer, Swin2021, chen2021crossvit,ali2021xcit}, at the cost of reducing receptive fields or non-linearity capacity. Other studies proposed hybrid networks by introducing convolution operations to ViTs~\citep{wang2022pvt, dai2021coatnet, halonet}, designing hybrid architectures to combine CNN with ViT modules~\citep{dai2021coatnet, botnet, container}.

\noindent\textbf{State Space Models and RNNs.} 
State space models (SSMs)~\citep{s4dgu2022, s4ndnguyen2022s4nd, selectives4wang2023selective}, which have the long-sequence modeling capacity with linear complexity, are also migrated from the natural language area (Mamba~\citep{mambagu2023mamba}). Visual SSMs were also designed by adapting the selective scan mechanism to 2-D images~\citep{zhu2024vision, liu2024vmamba}. 
Nevertheless, SSMs based on the selective scan mechanism suffer from limited parallelism, restricting their overall potential. 
Recent receptance weighted key value (RWKV) and RetNet models~\citep{RWKV23,retnetsun2023retentive} improved the parallelism while retaining the linear complexity. They combine the efficient parallelizable training of transformers with the efficient inference of RNNs, leveraging a linear attention mechanism and allowing formulation of the model as either a Transformer or an RNN, thus parallelizing computations during training and maintaining constant computational and memory complexity during inference. Despite the advantages, modeling a 2-D image as a sequence impairs interpretability.
%
%Many flavors of structured state space models sprang up with structures like complex-diagonal structure ~\citep{dssgupta2022, s4dgu2022}, multiple-input multiple output supporting ~\citep{s5smith2022simplified}, decomposition of diagonal plus low-rank operations ~\citep{liquids4hasani2022liquid}, selection mechanism ~\citep{mambagu2023mamba}. These models are then integrated into large representation models ~\citep{gssmehta2023long, megama2022mega, h3fu2022hungry}. These SSMs are mainly focuses on the application of long-range and casual data like language and speech, such as language understanding~\citep{megama2022mega, gssmehta2023long},  content-based reasoning~\citep{mambagu2023mamba},  pixel-level 1-D image classification ~\citep{s4gu2022}. 

\noindent\textbf{Biology and Physics Inspired Models.}
Biology and physics principles have long been the fountainhead of creating vision models. 
Diffusion models~\citep{song2020denoising, ho2020denoising, saharia2022photorealistic}, motivated by Nonequilibrium thermodynamics~\citep{de2013non}, are endowed with the ability to generate images by defining a Markov chain for the diffusion step. 
QB-Heat~\citep{chen2022self} utilizes physical heat equation as supervision signal for masked image modeling task. Spiking Neural Network (SNNs)~\citep{ghosh2009spiking, tavanaei2019deep, lee2016training} claims better simulation on the information transmission of biological neurons, formulating models for simple visual tasks~\citep{bawane2018object}. 
%
%Nevertheless, these models lack the potential as foundation backbones for generic visual recognition.
%
The success of these models encourages us to explore the principle of physical heat conduction for the development of vision representation models.

\section{Methodology}

%To derive vHeat, we first review the physical heat equation and its general solution. Based on the general solution, we propose the heat conduction operator, which is plugged to the deep learning framework to create vHeat. A discussion is also provided, delivering new insights about this heat conduction inspired vision model.

\subsection{Preliminaries: Physical Heat Conduction}\label{heat equation}

%Let $(x,y)$ be a point with a two-dimensional region $\boldsymbol{D}\in\mathbb{R}^2$. Let $u(x,y,t)$ denote the temperature at point $(x,y)$ and time $t$. The classic physical heat equation~\citep{widder1976heat} is defined as
Let $u(x,y,t)$ denote the temperature of point $(x,y)$ at time $t$ within a two-dimensional region $D\in\mathbb{R}^2$, the classic physical heat equation~\citep{widder1976heat} can be formulated as
\begin{equation}\label{eq:heat}
    \frac{\partial{u}}{\partial{t}}=k\left(\frac{\partial^2{u}}{\partial{x^2}}+\frac{\partial^2{u}}{\partial{y^2}}\right),
\end{equation}
where $k>0$ is the \textbf{thermal diffusivity}~\citep{bird2002transport}, measuring the rate of heat transfer in a material. 
By setting the initial condition $u(x,y,t)|_{t=0}$ to $f(x,y)$, the general solution of Eq.~\eqref{eq:heat} can be derived by applying the Fourier Transform (FT, denoted as $\mathcal{F}$) to both sides of the equation, which gives
\begin{equation}\label{eq:ft}
    \mathcal{F} \left(\frac{\partial{u}}{\partial{t}} \right)=k\mathcal{F} \left(\frac{\partial^2{u}}{\partial{x^2}}+\frac{\partial^2{u}}{\partial{y^2}}\right).
\end{equation}

%Supposing the initial condition is
%\begin{equation}
%u(x,y,0)=f(x,y),
%\label{eq:init}
%\end{equation}
% we can use the Fourier Transform to calculate the general solution for $u(x,y,t)$. Specifically, Fourier Transform is applied at both sides of Eq.~\eqref{eq:heat} to have the following equation
% https://links.uwaterloo.ca/amath353docs/set10.pdf
% Performing FT at both sides of Eq.\eqref{eq:heat}, therefore we have:

Denoting $\widetilde{u}(\omega_x,\omega_y,t)$ as the FT-transformed form of $u(x,y,t)$, \textit{i.e.}, $\widetilde{u}(\omega_x,\omega_y,t) \coloneqq \mathcal{F}(u(x,y,t))$, the left-hand-side of Eq.~\eqref{eq:ft} can be written as
\begin{equation}\label{eq:LHS}
    \mathcal{F} \left(\frac{\partial{u}}{\partial{t}} \right)=\frac{\partial{\widetilde{u}(\omega_x,\omega_y,t)}}{\partial{t}}.
\end{equation}
and by leveraging the derivative property of FT, the right-hand-side of Eq.~\eqref{eq:ft} can be transformed as
\begin{equation}\label{eq:RHS}
    \mathcal{F} \left(\frac{\partial^2{u}}{\partial{x^2}}+\frac{\partial^2{u}}{\partial{y^2}}\right)=-(\omega_x^2+\omega_y^2)\widetilde{u}(\omega_x,\omega_y,t).
\end{equation}
Therefore, by combining the expression of both sides of the equation, Eq.~\eqref{eq:ft} can be formulated as an ordinary differential equation (ODE) in the frequency domain, which can be written as
\begin{equation}\label{eq:ode}
    \frac{d{\widetilde{u}(\omega_x,\omega_y,t)}}{dt}=-k(\omega_x^2+\omega_y^2)\widetilde{u}(\omega_x,\omega_y,t).
\end{equation}
By setting the initial condition $\widetilde{u}(\omega_x,\omega_y,t)|_{t=0}$ to $\widetilde{f}(\omega_x,\omega_y)$ ($\widetilde{f}(\omega_x,\omega_y)$ denotes the FT-transformed $f(x,y)$), $\widetilde{u}(\omega_x,\omega_y,t)$ in Eq~\eqref{eq:ode} can be solved as
\begin{equation}\label{eq:solution_freq}
    \widetilde{u}(\omega_x,\omega_y,t)=\widetilde{f}(\omega_x,\omega_y)e^{-k(\omega_x^2+\omega_y^2)t}.
\end{equation}
Finally, the general solution of heat equation in the spatial domain can be obtained by performing inverse Fourier Transformer ($\mathcal{F}^{-1}$) on Eq.~\eqref{eq:solution_freq}, which gives the following expression
\begin{align}\label{eq:solution_spatial}
    u(x,y,t)&=\mathcal{F}^{-1}(\widetilde{f}(\omega_x,\omega_y)e^{-k(\omega_x^2+\omega_y^2)t}). %\\
            %&=\frac{1}{4\pi^2}{\int_{\widetilde{D}}}\widetilde{f}(\omega_x,\omega_y)e^{-k(\omega_x^2+\omega_y^2)t}e^{i({\omega_x}x+{\omega_y}y)}d{\omega_x}d{\omega_y}.
\end{align}

% By multiplying $\frac{dt}{\widetilde{u}(\omega_x,\omega_y,t)}$ to both sides of Eq.~\eqref{eq:ode}, we have
%
% \begin{equation}
% \frac{d{\widetilde{u}(\omega_x,\omega_y,t)}}{\widetilde{u}(\omega_x,\omega_y,t)}=-k(\omega_x^2+\omega_y^2)dt.
% \label{eq:ode2}
% \end{equation}
% By integrating both sides of Eq.~\eqref{eq:ode}, we have
% \begin{equation}
% {\ln}\widetilde{u}(\omega_x,\omega_y,t)-{\ln}\widetilde{u}(\omega_x,\omega_y,0)=-k(\omega_x^2+\omega_y^2)t.
% \label{eq:ode3}
% \end{equation}
% Let $\widetilde{u}(\omega_x,\omega_y,0)=\widetilde{f}(\omega_x,\omega_y)$ denote the initial condition, where $\widetilde{f}(\omega_x,\omega_y)$ represents the Fourier transformed function of $f(x,y)$. Given the initial condition, $\widetilde{u}(\omega_x,\omega_y,t)$ can be calculated as

%Notably, this general solution is applicable to 2D heat conduction, which is fundamental for the subsequent proposed heat conduction operator.

\begin{figure}[t]
%\centering
%\begin{center}
\includegraphics[width=0.465\textwidth]{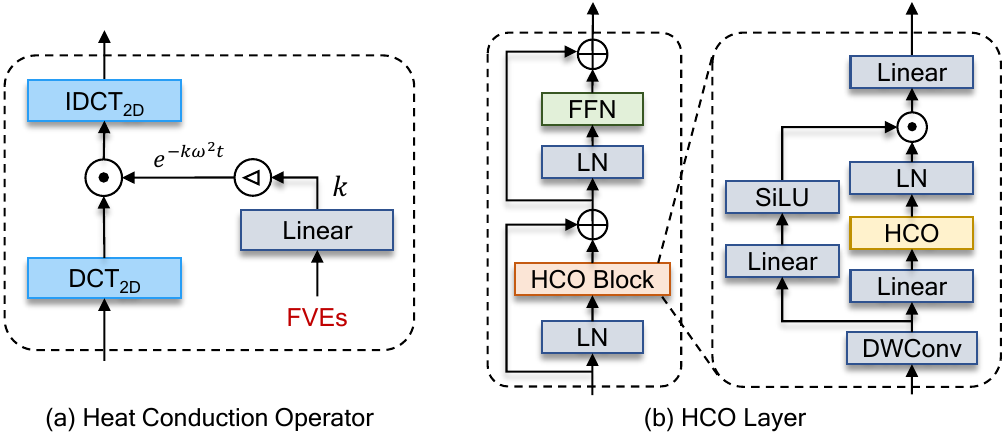}
\caption{HCO and HCO layer. FVEs, FFN, LN, and DWConv respectively denote frequency value embeddings, feed-forward network, layer normalization, and depth-wise convolution. Please refer to Sec. \ref{appendix: dwconv} in the supplementary, where we demonstrate that while depth-wise convolution aids in feature extraction, the primary improvements are attributed to the proposed HCO.}
\label{fig:hco}
%\end{center}
\end{figure}

%%%%%%%%%%%%%%%%%%%%%%%%%%%%%%%%%%%%%%%%%%%%%%%%%%%%%%%%%%%%%%%
\subsection{vHeat: Visual Heat Conduction}

%In this subsection, we introduce the heat conduction operator and vHeat models, covering aspects such as discretizating the heat conduction equation for integration into deep models, optimizing HCO by matrix calculation for GPUs, and examining the impact of thermal diffusivity, \textit{etc}.

Drawing inspiration from the analogies between the principles of physical heat conduction and the propagation of visual semantics within the spatial domain (\textit{i.e.}, `visual heat conduction'), we propose \textbf{vHeat}, a physics-inspired deep architecture for visual representation learning.
The vHeat model is built upon the Heat Conduction Operator (HCO), which is designed to integrate the principle of heat conduction into handling the discrete feature of vision data.
We also leverage the thermal diffusivity in the classic physical heat equation (Eq~\eqref{eq:heat}) to improve the adaptability of vHeat to vision data.

\subsubsection{Heat Conduction Operator (HCO)}

To extract visual features, we design HCO to implement the conduction of visual information across image patches in multiple dimensions, following the principle of physical heat conduction. 
To this end, we first extend the 2D temperature distribution $u(x,y,t)$ along the channel dimension and denote the resultant multi-channel image feature as $U(x,y,c,t)$ ($c=1,\cdots,C$).
Mathematically, considering the input as $U(x,y,c,0)$ and the output as $U(x,y,c,t)$, HCO simulates the general solution of physical heat conduction (Eq.~\eqref{eq:solution_spatial}) in visual data processing, which can be formulated as 
\begin{equation}\label{eq:solution_U}
    U^t=\mathcal{F}^{-1}(\mathcal{F}(U^0)e^{-k(\omega_x^2+\omega_y^2)t}),
\end{equation}
where $U^t$ and $U^0$ are abbreviations for $U(x,y,c,t)$ and $U(x,y,c,0)$, respectively.

% Let $U^t=U(x,y,c,t)$ and $U^0=U(x,y,c,0)$ for short. 
%
% According to the general solution defined by Eq.~\eqref{eq:solution_spatial}, HCO is formulated as
% \begin{equation}
%     U^t=\mathcal{F}^{-1}(\mathcal{F}(U^0)e^{-k(\omega_x^2+\omega_y^2)t}).
% \label{eq:solution_U}
% \end{equation}

For applying $\mathcal{F}(\cdot)$ and $\mathcal{F}^{-1}(\cdot)$ to discrete image patch features, it is necessary to utilize the discrete version of the (inverse) Fourier Transform (\textit{i.e.}, DFT and IDFT). 
However, since vision data is spatially constrained and semantic information will not propagate beyond the border, we additionally introduce a common assumption of Neumann boundary condition~\citep{cheng2005heritage}, \textit{i.e.}, $\partial{u(x,y,t)}/\partial{\mathbf{n}}=0, \forall{(x,y)\in{\partial}D,t\geq0}$, where $\mathbf{n}$ denotes the normal to the image boundary $\partial{D}$. 
%
% Take a further step, we utilize a common assumption of Neumann boundary condition~\citep{cheng2005heritage}, $\frac{\partial{u(x,y,t)}}{\partial{\mathbf{n}}}=0, \forall{(x,y)\in{\partial}D,t\geq0}$, where $\mathbf{n}$ denotes the (typically exterior) normal to the boundary $\partial{D}$. 
%
As vision data is typically rectangular, this boundary condition enables us to replace the 2D DFT and IDFT with the 2D discrete cosine transformation, $\mathbf{DCT_{2D}}$, and the 2D inverse discrete cosine transformation, $\mathbf{IDCT_{2D}}$~\citep{strang1999discrete}. 
Therefore, the discrete implementation of HCO can be expressed as
\begin{equation}\label{eq:implementation_U}
    U^t=\mathbf{IDCT_{2D}}(\mathbf{DCT_{2D}}(U^0)e^{-k(\omega_x^2+\omega_y^2)t}),
\end{equation}
and its internal structure is illustrated in Fig.~\ref{fig:hco}(a). 
Particularly, the parameter $k$ stands for the thermal diffusivity in physical heat conduction and is predicted based on the features within the frequency domain (explained in the following subsection). 

Notably, due to the computational efficiency of $\mathbf{DCT_{2D}}$, the overall complexity of HCO is $\mathcal{O}(N^{1.5})$, where $N$ denotes the number of input image patches.
Please refer to Sec.~\ref{appendix: DCT} in the supplementary for the detailed implementation of HCO using $\mathbf{DCT_{2D}}$ and $\mathbf{IDCT_{2D}}$. 

\begin{figure}[t]
\centering
\begin{center}
\includegraphics[width=0.49\textwidth]{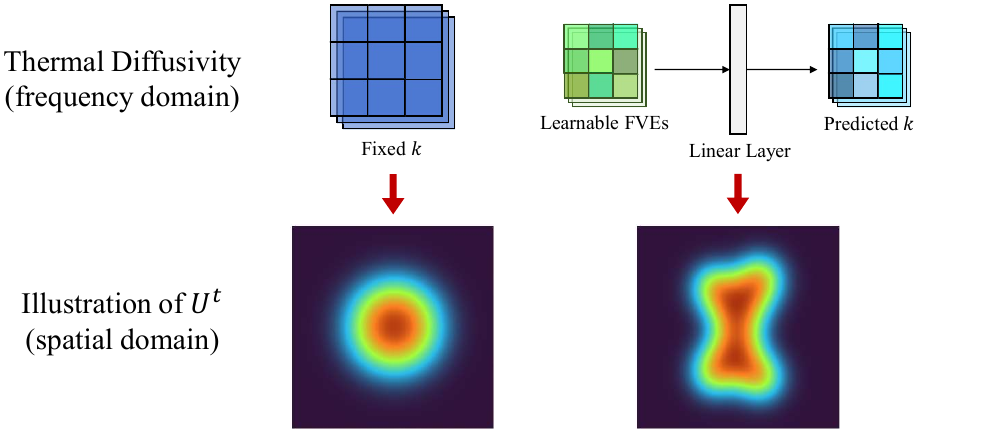} %k1有底板配色，k2没有，可以替换
\caption{Illustration of temperature distribution $U^t$ $w.r.t.$ thermal diffusivity $k$, given a heat source as the initial condition. The predicted $k$ leads to nonuniform visual heat conduction, which facilitates the adaptability of visual representation. (Best viewed in color)} %预测k使得热传导可以非均匀
\label{fig:k}
\end{center}
\vspace{-20pt}
\end{figure}

\subsubsection{Adaptive Thermal Diffusivity}

% 在物理学中，热扩散率表示某种材料热量传递的速度，即可以guide温度分布。对应地，对于图像来说，热扩散率应该表示的是某个像素点（或patch）传递其自身信息（热量）的速度。由于图像信息是不规则的，一个合理的假设是图像的热扩散也应该是不规则的，例如重要的像素点（或patch）应该有更快更远的扩散能力 -- 图像数据应该具有自适应的热扩散属性。
% 对图像内容来说，像素点（或patch）的重要性可以由其对应的频率表示，这个claim是make sence的，正如已有的研究工作deliver的那样：高频的信息往往可以代表图像的重要属性~\citep{}。
% 所以，图像数据的自适应热传导可以在频域上实现。
% 进一步地，为了实现频域的自适应性，如公式12/13所示，我们对傅里叶变换得到的频域（$\mathbf{DCT_{2D}}(U^0)$）执行自适应滤波操作，即$\mathbf{DCT_{2D}}(U^0)e^{-k(\omega_x^2+\omega_y^2)t}$. 其中$\omega_x$和$\omega_y$是标量, 为了实现自适应性，我们将k设计为可学习的变量，从而实现自适应的滤波过程，进而实现对图像频域的自适应。 
% （然后结合画出的图做一些分析...）

% 另外，DCT之后的频域值是与位置相关的，但是经过DCT变换之后的特征map损失了位置信息。为了保留在频域上的位置信息，我们将k升级为可学习的learnable frequency position embeddings (FVEs)。具体地，FVEs are implemented by the popularly utilized absolute position embeddings in ViTs, which are then followed by a linear layer to yield k.
% 原则上来说，传导时间t也影响了滤波过程，但是考虑到k和t是相乘的关系，所以我们固定t为标量1. 

%讲一下传热系数及其作用
In physical heat conduction, thermal diffusivity represents the rate of heat transfer within a material. 
While in visual heat conduction, we hypothesize that more representative image contents contain more energy, resulting in higher temperatures in the corresponding image features within $U(x,y,c,t)$. 
Therefore, it is suggested that the thermal diffusivity parameter $k$ should be learnable and adaptive to image content, which facilitates the adaptability of heat condution to visual representation learning.

Given that the output of $\mathbf{DCT}$ (i.e., $\mathbf{DCT_{2D}}(U^0)$ in Eq.~\eqref{eq:implementation_U}) lies in the frequency domain, we also determine $k$ based on frequency values ($k \coloneqq k(\omega_x,\omega_y)$). 
Since different positions in the frequency domain correspond to different frequency values, we propose to represent these values using learnable Frequency Value Embeddings (FVEs), which function similarly to the widely used absolute position embeddings in ViTs\cite{ViT2021} (despite in the frequency domain).
As shown in Figure~\ref{fig:hco} (a), FVEs are fed to a linear layer to predict the thermal diffusivity $k$, allowing it to be non-uniform and adaptable to visual representations.

Practically, considering that $k$ and $t$ (the conduction time) are multiplied in Eq.~\eqref{eq:implementation_U}, we empirically set a fixed value for $t$ and predict the values of $k$. 
Specifically, FVEs are shared within each network stage of vHeat to facilitate the convergence of the training process.

%However, in contrast to architectures based on CNN or ViT, which organize features into spatial maps expanded along the channel dimension, representations acquired through HCO are in the frequency domain.

% To fulfill this purpose, considering $\mathbf{DCT_{2D}}(U^0)$ is in the frequency domain, we also predict $k$ Eq.~\eqref{eq:implementation_U} upon frequency values, $k \coloneqq k(\omega_x,\omega_y)$. 
%
% After $\mathbf{DCT}$, different positions in the frequency domain represent the information of different frequency values. To retain it, we propose the learnable frequency value embeddings (FVEs) to represent the frequency values. 
%
% FVEs are implemented the same as the widely utilized absolute position embeddings in ViTs~\citep{ViT2021}, which are fed to a linear layer to predict the thermal diffusivity $k$, Fig.~\ref{fig:k} (Right). 
%
% With the predicted $k$, visual heat conduction could be nonuniform, which facilitates the adaptability of visual representation. 
%
% Considering the thermal diffusivity $k$ and the conduction time $t$ is multiplied in Eq.~\eqref{eq:implementation_U}, we empirically set a fixed value for $t$ and predict the values of $k$. In particular, we share FVEs in each stage to facilitate an easier training process. 

\subsubsection{vHeat Model}

\noindent\textbf{Network Architecture.}
% As shown in Fig.~\ref{fig:vheat_arch}, vHeat follows a hierarchical deep network architecture with gradually decreased resolutions of $\frac{H}{4}\times \frac{W}{4}$, $\frac{H}{8}\times \frac{W}{8}$, $\frac{H}{16}\times \frac{W}{16}$, and $\frac{H}{32}\times \frac{W}{32}$ and increasing channels ~\citep{Resnet2016, Swin2021}, where $H$ and $W$ represent the height and the width of the input image, respectively.
% %
% In vHeat, multiple stages are stacked, each composed of a down-sampling layer followed by multiple heat conduction layers (except for the initial stage). 
% %
% We introduce a vHeat model family including vHeat-Tiny (vHeat-T), vHeat-Small (vHeat-S), and vHeat-Base (vHeat-B). 
%The large-scale model will be explored in the future.
%
We develop a vHeat model family including vHeat-Tiny (vHeat-T), vHeat-Small (vHeat-S), and vHeat-Base (vHeat-B). An overview of the network architecture of vHeat is illustrated in Fig.~\ref{fig:vheat_arch}, and the detailed configurations are provided in Sec. \ref{appendix: settings} in Appendix. 
Given an input image with the spatial resolution of $H\times W$, vHeat first partitions it to image patches through a stem module, yielding a 2D feature map with $\frac{H}{4}\times \frac{W}{4}$ resolution. 
Subsequently, multiple stages are utilized to create hierarchical representations with gradually decreased resolutions of $\frac{H}{4}\times \frac{W}{4}$, $\frac{H}{8}\times \frac{W}{8}$, $\frac{H}{16}\times \frac{W}{16}$ and increasing channels. 
Each stage is composed of a down-sampling layer followed by multiple heat conduction layers (except for the first stage).

%%%%%%%%%%%%%%%%%%%%%%%%%%%%%%%%%%%%%%%%%%%%%%%%%%%%%%%%%%%%%%%%%
%这里需要对照Fig. 2(b)陈述更多HCO layer实现的细节，对一些重要的模块予以说明
%%%%%%%%%%%%%%%%%%%%%%%%%%%%%%%%%%%%%%%%%%%%%%%%%%%%%%%%%%%%%%%%%
\noindent\textbf{Heat Conduction Layer.}
The heat conduction layer, Fig.~\ref{fig:hco} (b), is similar to the ViTs block while replacing self-attention operators with HCOs and retaining the feed-forward network (FFN). 
It first utilizes a $3\times3$ depth-wise convolution layer. 
%\footnote{In experiment, we will demonstrate that while depth-wise convolution aids in feature extraction, the main gains come from the proposed HCO to extract the feature of the input, 
The depth-wise convolution is followed by two branches: one maps the input to HCO and the other computes the multiplicative gating signal like ~\citep{liu2024vmamba}. 
HCO plays a crucial role in each heat conduction layer, Fig.~\ref{fig:hco} (b), where the mapped features from a linear layer are first processed by the $\mathbf{DCT_{2D}}$ operator to generate features in the frequency domain.
%
%{\color{red}
%HCO feeds a frequency input, which is implemented using the frequency position %embeddings namely FVEs, Fig.~\ref{fig:hco} (a)). 
%}
Additionally, HCO takes FVEs as input for frequency representation to predict adaptive thermal diffusivity $k$ through a linear layer. 
%
%The FVEs are followed by a linear layer, yielding the thermal diffusivity $k$, which is then converted to a thermal diffusivity coefficient ($e^{-kw^2t}$) related to the features in the frequency domain.
%
%Multiplying the coefficient matrix $e^{-k{\omega^2}t}$ with the $\mathbf{DCT}$ features implements a filtering procedure, which selects the most informative and discriminative features in the frequency domain.
%
By multiplying the coefficient matrix $e^{-k{\omega^2}t}$ and performing $\mathbf{IDCT_{2D}}$, HCO implements the discrete solution of the visual heat equation, Eq.~\eqref{eq:implementation_U}. 
%
%By processing $\mathbf{IDCT_{2D}}$ to the multiplied features, HCO outputs visual heat conduction
%
%Since the Heat2D module fundamentally operates on 2D features and maintains the token positions, we do not employ the position embedding bias.

\begin{table*}[ht]
\centering
% \small
% \addtolength{\tabcolsep}{-5.pt}
\setlength{\tabcolsep}{0.25cm}
\caption{Performance comparison of image classification on ImageNet-1K. Test throughput values are measured with an A100 GPU, using the toolkit released by~\citep{rw2019timm}, following the protocol proposed in~\citep{Swin2021}. The batch size is set as 128, and the PyTorch version is 2.2. % and ablation results compared with GFNet~\citep{gfnet}. 
%Throughput values are measured with an A100 GPU and an AMD EPYC 7542 CPU, using the toolkit released by~\citep{rw2019timm}, following the protocol proposed in~\citep{Swin2021}.
}
\scalebox{0.95}{
\begin{tabular}{c|cccc|c}
\toprule
Method & \begin{tabular}[c]{@{}c@{}}Image \\ size\end{tabular} & \#Param. & FLOPs & \begin{tabular}[c]{@{}c@{}}Test Throughput\\(img/s)\end{tabular} & \begin{tabular}[c]{@{}c@{}}ImageNet \\ top-1 acc.  (\%)\end{tabular} \\
%\midrule
%ViT-B/16~\citep{ViT2021} & 384$^2$ & 86M & 55.4G & - & 77.9 \\
%ViT-L/16~\citep{ViT2021} & 384$^2$ & 307M & 190.7G & - & 76.5 \\
% \midrule
% DeiT-S~\citep{DeiT2021} & 224$^2$ & 22M & 4.6G & 1761 & 79.8 \\
% DeiT-B~\citep{DeiT2021} & 224$^2$ & 86M & 17.5G & 503 & 81.8 \\
%DeiT-B~\citep{DeiT2021} & 384$^2$ & 86M & 55.4G & - & 83.1 \\
\midrule
Swin-T~\citep{Swin2021} & 224$^2$ & 28M & 4.6G & 1242 & 81.3 \\
ConvNeXt-T~\citep{liu2022convnet} & 224$^2$ & 29M & \textbf{4.5G} & 1198 & 82.1 \\
DCFormer-SW-T~\citep{Li_2023_WACV} & 512$^2$ & 28M & \textbf{4.5G} & - & 82.1 \\
Vim-S~\citep{zhu2024vision} & 224$^2$ & \textbf{26M} & 5.3G & 811 & 81.4 \\
%\rowcolor{Gray}
\rblue
vHeat-T (Ours) & 224$^2$ & 29M & 4.6G & \textbf{1514} & \textbf{82.2} \\
\midrule
Swin-S~\citep{Swin2021} & 224$^2$ & 50M & 8.7G & 720 & 83.0 \\
ConvNeXt-S~\citep{liu2022convnet} & 224$^2$ & 50M & 8.7G & 687 & 83.1 \\
DCFormer-SW-S~\citep{Li_2023_WACV} & 512$^2$ & 50M & 8.7G & - & 82.9 \\
%\rowcolor{Gray}
\rblue
vHeat-S (Ours) & 224$^2$ & 50M & \textbf{8.5G} & \textbf{945} & \textbf{83.6} \\
\midrule
Swin-B~\citep{Swin2021} & 224$^2$ & 88M & 15.4G & 456 & 83.5 \\
ConvNeXt-B~\citep{liu2022convnet} & 224$^2$ & 89M & 15.4G & 439 & 83.8 \\
RepLKNet-31B~\citep{replknet} & 224$^2$ & 79M & 15.3G & - & 83.5 \\
DCFormer-SW-B~\citep{Li_2023_WACV} & 512$^2$ & 88M & 15.4G & - & 83.5 \\
Vim-B~\citep{zhu2024vision} & 224$^2$ & 98M & 19.0G & 294 & 83.2 \\
%\rowcolor{Gray}
\rblue
vHeat-B (Ours) & 224$^2$ & \textbf{68M} & \textbf{11.2G} & \textbf{661} & \textbf{84.0} \\
\bottomrule
\end{tabular}
}
\normalsize
\label{tab:imagenet}
\vspace{-10pt}
\end{table*}

\subsection{Discussion }
%讨论一下为什么物理传热能够被用于建模视觉模型？

\noindent$\bullet$\hspace{0.1cm}\noindent\textbf{What is role of the thermal diffusivity coefficient $e^{-k(\omega_x^2+\omega_y^2)t}$?} 
When multiplying with $\mathbf{DCT_{2D}}(U^0)$, $e^{-k(\omega_x^2+\omega_y^2)t}$ acts as an adaptive filter in the frequency domain to perform visual heat conduction. 
Different frequency values correspond to distinct image patterns, $i.e.,$ high frequency corresponds to edges and textures while low frequency corresponds to flat regions. 
With adaptive thermal diffusivity, HCO can enhance/depress these patterns within each feature channel. Aggregating the filtered features from all channels, vHeat achieves a robust feature representation.
%Consequently, adaptive thermal diffusivity in image data can be achieved in the frequency domain. To this end, we perform adaptive filtering on the frequency domain obtained from the Fourier transform ($\mathbf{DCT_{2D}}(U^0)$), expressed as $\mathbf{DCT_{2D}}(U^0)e^{-k(\omega_x^2+\omega_y^2)t}$. Here, $\omega_x$ and $\omega_y$ are scalars, and to ensure adaptivity, we design $k$ as a learnable variable.
% 在物理学中，热扩散率表示某种材料热量传递的速度，即可以guide温度分布。对应地，对于图像来说，热扩散率应该表示的是某个像素点（或patch）传递其自身信息（热量）的速度。由于图像信息是不规则的，一个合理的假设是图像的热扩散也应该是不规则的，例如重要的像素点（或patch）应该有更快更远的扩散能力 -- 图像数据应该具有自适应的热扩散属性。
% 对图像内容来说，像素点（或patch）的重要性可以由其对应的频率表示，这个claim是make sence的，正如已有的研究工作deliver的那样：高频的信息往往可以代表图像的重要属性~\citep{}。
% 所以，图像数据的自适应热传导可以在频域上实现。
% 进一步地，为了实现频域的自适应性，如公式12/13所示，我们对傅里叶变换得到的频域（$\mathbf{DCT_{2D}}(U^0)$）执行自适应滤波操作，即$\mathbf{DCT_{2D}}(U^0)e^{-k(\omega_x^2+\omega_y^2)t}$. 其中$\omega_x$和$\omega_y$是标量, 为了实现自适应性，我们将k设计为可学习的变量，从而实现自适应的滤波过程，进而实现对图像频域的自适应。 

% $\bullet$\hspace{0.1cm}\noindent\textbf{Why does higher temperature $U(x,y,c,t)$ correspond to more representative features?}
% %
% Higher temperatures indicate more information convergence, which implies a higher correlation of image patches and more informative features. 
% %
% This is similar to the self-attention operation, which uses the correlation coefficients between patches to select representative features. 
% %
% %Heat conduction shares this essence of information conduction, achieved through temperature as represented by $U(x,y,c,t)$.

\noindent$\bullet$\hspace{0.1cm}\noindent\textbf{Why does temperature $U(x,y,c,t)$ correspond to visual features?}
% %
% Higher temperatures indicate more information convergence, which implies a higher correlation of image patches and more informative features. 
% %
% This is similar to the self-attention operation, which uses the correlation coefficients between patches to select representative features. 
% %
Visual features are essentially the outcome of the feature extraction process, characterized by pixel propagation within the feature map. This process aligns with the properties of existing convolution, self-attention, and selective scan operators, exemplifying a form of information conduction. Similarly, visual heat conduction embodies this concept of information conduction through temperature, denoted as $U(x,y,c,t)$. 
%Visual features are essentially the result of the feature extraction process, characterized by pixel propagation within the feature map. This aligns with the properties of existing convolution, self-attention, and selective scan operators. This characteristic is a form of information conduction. Heat conduction shares this essence of information conduction, achieved through temperature, \textit{i.e.}, $U(x,y,c,t)$.

% 参考我之前写的footnote: The convolutional, attention, and selective scan operators all facilitate the characteristics of pixel propagation within the feature map, namely information conduction. 
% Representive features实际上是某种特征提取的结果，这种特征提取过程具有pixel propagation within the feature map的特点，这也符合现存的convolutional, attention, and selective scan operators的特性。而这一特点实际上就是一种信息传导。heat conduction的本质正是一种信息传导，只不过这种传导是通过temperature实现的，即$U(x,y,c,t)$。

\noindent$\bullet$\hspace{0.1cm}\noindent\textbf{What is the relationship/difference between HCO and self-attention?} HCO dynamically propagates energy via heat conduction, enabling the perception of global information within the input image. This positions HCO as a distinctive form of attention mechanism. The distinction lies in its reliance on interpretable physical heat conduction, in contrast to self-attention, which is formulated through token similarity. Furthermore, HCO works in the frequency domain, implying its potential to affect all image patches through frequency filtering. Consequently, HCO exhibits greater efficiency compared to self-attention, which necessitates computing the relevance of all pairs across image patches. 
%HCO adaptively propagates energy through heat conduction, thereby perceiving the global information of the input image. This means that HCO functions as a special kind of attention mechanism. The difference lies in that HCO is based on the interpretable physical heat conduction model, while self-attention is formulated through the similarity. Besides, HCO actually works in the frequency domain, which implies that HCO could have impacts on all image patches by frequency filtering. Therefore, HCO is more efficient than self-attention, which has to calculate the relevance of all pairs among all image patches. 
\section{Experiment \& Analysis}

Experiments are performed to assess vHeat and compare it against popular CNN and ViT models. Visualization analysis is presented to gain deeper insights into the mechanism of vHeat. The evaluation spans image classification, object detection, semantic segmentation, out-of-distribution classification, and low-level vision tasks. Please refer to Sec. \ref{appendix: settings} in the supplementary for experimental settings.  

\subsection{Experimental Results}

%\subsection{Image Classification}
%
\noindent\textbf{Image classification.} The image classification results are summarized in Table ~\ref{tab:imagenet}. With similar FLOPs, vHeat-T achieves $82.2\%$ top-1 accuracy, outperforming Swin-T/Vim-S by $0.9\%/0.8\%$, respectively. 
Notably, the superiority of vHeat is also observed at both Small and Base scales. Specifically, vHeat-B achieves a top-1 accuracy of $84.0\%$ with only 11.2G FLOPs and 68M model parameters, outperforming Swin-B/Vim-B by $0.5\%/0.8\%$, respectively.

In terms of computational efficiency, vHeat enjoys significantly higher inference speed across Tiny/Small/Base model scales compared to benchmark models. For instance, vHeat-T achieves a throughput of 1514 images/s, $87\%$ higher than Vim-S, $26\%$ higher than ConvNeXt-T, and $22\%$ higher than Swin-T, while maintaining a performance superiority, respectively.

%\subsection{Downstream Task}
\begin{table}[h]
\setlength{\tabcolsep}{0.17cm}
\centering
\caption{Results of object detection and instance segmentation on COCO. FLOPs are calculated with input size $1280\times800$. $AP^{b}$ and $AP^{m}$ denote box AP and mask AP, respectively. The notation `$1\times$' indicates models fine-tuned for 12 epochs, while `$3\times$MS' denotes the utilization of multi-scale training for 36 epochs.}
% \scalebox{0.85}{
\begin{tabular}{c|c|c|cc}
    %\caption{Results of object detection and instance segmentation on COCO dataset. FLOPs are calculated with input size $1280\times800$. $AP^{b}$ and $AP^{m}$ denote box AP and mask AP, respectively. The notation `$1\times$' indicates models fine-tuned for 12 epochs, while `$3\times$MS' denotes the utilization of multi-scale training for 36 epochs.}
    \toprule
    \multicolumn{5}{c}{\textbf{Mask R-CNN 1$\times$ schedule on COCO}}\\
    \midrule
    Backbone & AP$^\text{b}$ & AP$^\text{m}$ & FPS (images/s) & FLOPs \\
    \midrule
    %ResNet-50 & 38.2 & 34.7 & 44M & 260G \\
    Swin-T & 42.7 & 39.3 & 26.3 & 267G \\
    ConvNeXt-T & 44.2 & 40.1 & 29.3 & \textbf{262G} \\
    %PVTv2-B2 & 45.3 & 67.1 & 49.6 & 41.2 & 64.2 & 44.4 & 45M & 309G \\
    %ViT-Adapter-S & 44.7 & 65.8 & 48.3 & 39.9 & 62.5 & 42.8 & 48M & 403G \\
    %\rowcolor{Gray}
    \rblue
    vHeat-T (Ours) & \textbf{45.1} & \textbf{41.2} & \textbf{32.7} & 272G \\
    \midrule
    %ResNet-101 & 38.2 & 34.7 & 63M & 336G \\
    Swin-S & 44.8 & 40.9 & 19.7 & 359G \\
    ConvNeXt-S & 45.4 & 41.8 & 20.2 & 349G \\
    % PVTv2-B3 & 47.0 & 68.1 & 51.7 & 42.5 & 65.7 & 45.7 & 65M & 397G \\
    %\rowcolor{Gray}
    \rblue
    vHeat-S (Ours) & \textbf{46.8} & \textbf{42.3} & \textbf{25.9} & \textbf{348G} \\
    \midrule
    %Swin-B & 46.9 & 42.3 & 107M & 496G \\
    Swin-B & 46.9 & 42.3 & 13.8 & 504G \\
    ConvNeXt-B & 47.0 & 42.7 & 14.1 & 486G \\
    % PVTv2-B5 & 47.4 & 68.6 & 51.9 & 42.5 & 65.7 & 46.0 & 102M & 557G \\
    %ViT-Adapter-B & 47.0 & 41.8 & 102M & 557G \\
    %\rowcolor{Gray}
    \rblue
    vHeat-B (Ours) & \textbf{47.7} & \textbf{43.0} & \textbf{20.2} & \textbf{432G} \\
    \midrule
    \multicolumn{5}{c}{\textbf{Mask R-CNN 3$\times$ MS schedule on COCO}}\\
    \midrule
    % ResNet-50 &  / & / & / & / & / & / & 44M & 260G \\
    Swin-T &  46.0 & 41.6 & 26.3 & 267G \\
    ConvNeXt-T &  46.2 & 41.7 & 29.3 & \textbf{262G} \\
    %PVTv2-B2 &  47.8 & 69.7 & 52.6 & 43.1 & 66.8 & 46.7 & 45M & 309G \\
    %ViT-Adapter-S &  48.2 & 69.7 & 52.5 & 42.8 & 66.4 & 45.9 & 48M & 403G \\
    %\rowcolor{Gray}
    \rblue
    vHeat-T (Ours) & \textbf{47.2} & \textbf{42.4} & \textbf{32.7} & 272G \\ % epoch 33
    \midrule
    Swin-S & 48.2 & 43.2 & 19.7 & 359G \\
    ConvNeXt-S & 47.9 & 42.9 & 20.2 & 349G \\
    %PVTv2-B3 & 48.4 & 43.2 & 65M & 397G \\
    %\rowcolor{Gray}
    \rblue
    vHeat-S (Ours) & \textbf{48.8} & \textbf{43.7} & \textbf{25.9} & \textbf{348G} \\
    \bottomrule
    \end{tabular}
    % }
\label{table:detection}
\vspace{-20pt}
\end{table}

\noindent\textbf{Object Detection and Instance Segmentation.}
As a backbone network, vHeat is tested on the MS COCO 2017 dataset~\citep{COCO2014} for object detection and instance segmentation. We load classification pre-trained vHeat weights for downstream evaluation. Considering the input image size is different from the classification task, the shape of FVEs or $k$ should be aligned to the target image size on downstream tasks. Please refer to Sec. \ref{appendix: interpolation} in the supplementary for ablation of interpolation for downstream tasks. 
The results for object detection are summarized in Table ~\ref{table:detection}, and vHeat enjoys superiority in box/mask Average Precision (AP$^b$ and AP$^m$) in both of the training schedules ($12$ or $36$ epochs). 
For example, with a $12$-epoch fine-tuning schedule, vHeat-T/S/B models achieve object detection mAPs of $45.1\%/46.8\%/47.7\%$, outperforming Swin-T/S/B by $2.4\%/2.0\%/0.8\%$ mAP, and ConvNeXt-T/S/B by $0.9\%/1.4\%/0.7\%$ mAP, respectively. 
With the same configuration, vHeat-T/S/B achieve instance segmentation mAPs of $41.2\%/42.3\%/43.0\%$, outperforming Swin-T/S/B and ConvNeXt-T/S/B. 
The advantages of vHeat persist under the $36$-epoch ($3\times$) fine-tuning schedule with multi-scale training. 
Besides, vHeat enjoys much higher inference speed (FPS) compared with Swin and ConvNeXt. For example, vHeat-B achieves \textbf{20.2} images/s, \textbf{46\%/43\%} higher than Swin-B/ConvNeXt-B (13.8/14.1 images/s). 
These results highlight vHeat's potential to deliver strong performance and efficiency in dense prediction downstream tasks.
%These results showcase vHeat's potential to achieve promising performance and efficiency in downstream tasks with dense prediction.

\begin{table}[t]
\setlength{\tabcolsep}{0.25cm}
\centering
\caption{Results of semantic segmentation on ADE20K using UperNet~\citep{upernet}. FLOPs are calculated with the input size of $512\times512$.}
% \scalebox{0.87}{
    \begin{tabular}{c|c|cc}
    %\caption{Results of semantic segmentation on ADE20K using UperNet~\citep{upernet}. FLOPs are calculated with input size of $512\times2048$.}
    \toprule
    \multicolumn{4}{c}{\textbf{UperNet on ADE20K}}\\
    \midrule
    Backbone & mIoU & FPS (images/s) & FLOPs \\
    \midrule
    %ResNet-50 & 42.1 & 67M & 953G  \\
    %DeiT-S + MLN & 43.8 & 58M & 1217G \\
    Swin-T & 44.4 & 31.8 & 237G \\
    ConvNeXt-T & 46.0 & \textbf{37.8} & \textbf{235G} \\
    ViL-S & 46.3 & - & - \\
    %\rowcolor{Gray}
    \rblue
    vHeat-T (Ours) & \textbf{46.9} & 36.7 & \textbf{235G} \\
    \midrule
    %ResNet-101 & 43.8 & 86M & 1030G  \\
    %DeiT-B + MLN & 45.5 & 144M & 2007G \\
    Swin-S & 47.6 & 22.1 & 261G \\
    NAT-S & 48.0 & 23.1 & \textbf{254G} \\
    ConvNeXt-S & 48.7 & \textbf{27.7} & 257G \\
    %\rowcolor{Gray}
    \rblue
    vHeat-S (Ours) & \textbf{49.1} & 26.1 & \textbf{254G} \\
    \midrule
    Swin-B & 48.1 & 19.2 & 299G \\
    NAT-B & 48.5 & 20.8 & \textbf{285G} \\
    ViL-B & 48.8 & - & - \\
    ConvNeXt-B & 49.1 & 21.6 & 293G \\
    %\rowcolor{Gray}
    \rblue
    vHeat-B (Ours) & \textbf{49.6} & \textbf{23.6} & 293G  \\
    % \midrule
    % Swin-S & $640^{2}$ & 47.9 & 48.8 & 81M & 1614G \\
    % ConvNeXt-S & $640^{2}$ & 48.8 & 48.9 & 82M & 1607G \\
    % \rowcolor{Gray}
    % VMamba-S & $640^{2}$ & 50.8 & 50.8 & 76M & 1620G  \\ % iter112000
    \bottomrule
    \end{tabular}
    % }
\label{tab:segmentation}
%\vspace{-10pt}
\end{table}

\begin{table}
\setlength{\tabcolsep}{0.3cm}
\centering
    \caption{Robust comparison of vHeat-B with Swin-B.}
    %\vspace{-15pt}
    % \scalebox{0.9}{
    \begin{tabular}{l|ccccc}
        \toprule
        \multirow{2}{*}{Model} & ObjectNet & ImageNet-A \\ 
         & top-1 acc. (\%)  & top-1 acc. (\%)  \\
        \midrule
        Swin-B & 25.4 & 36.0 \\
        ConvNeXt-B & 26.1 & 36.5 \\
        %\rowcolor{Gray}
        \rblue
        vHeat-B (Ours) & \textbf{26.7} & \textbf{36.8} \\
        %ViT-B & 26.2 & 35.7 & 49.3 & 36.4 & 49.7 \\
        \bottomrule
    \end{tabular}
    % }
    \label{tab: robust}
\end{table}

\noindent\textbf{Semantic Segmentation.}
The results on ADE20K are summarized in Table ~\ref{tab:segmentation}, and vHeat consistently achieves superior performance over other baseline models across Tiny/Small/Base scales. For example, vHeat-B respectively outperform NAT-B~\citep{nat} and ViL-B~\citep{visionxlstm} by 1.1\%/0.8\% mIoU.

\noindent\textbf{Robustness evaluation.} To validate the robustness of vHeat, We evaluated vHeat-B on out-of-distribution classification datasets, including ObjectNet~\citep{objectnet} and ImageNet-A~\citep{imageneta}. We measure the Top-1 accuracy (\%) for these two benchmarks, Table ~\ref{tab: robust}. It is evident that vHeat outperforms Swin and ConvNeXt consistently (better results are marked in bold). These experiments highlight vHeat's robustness across out-of-distribution data, such as rotated objects, different view angles (ObjectNet), and natural adversarial examples (ImageNet-A).

\begin{figure}[t]
\centering
\begin{center}
\includegraphics[width=0.48\textwidth]{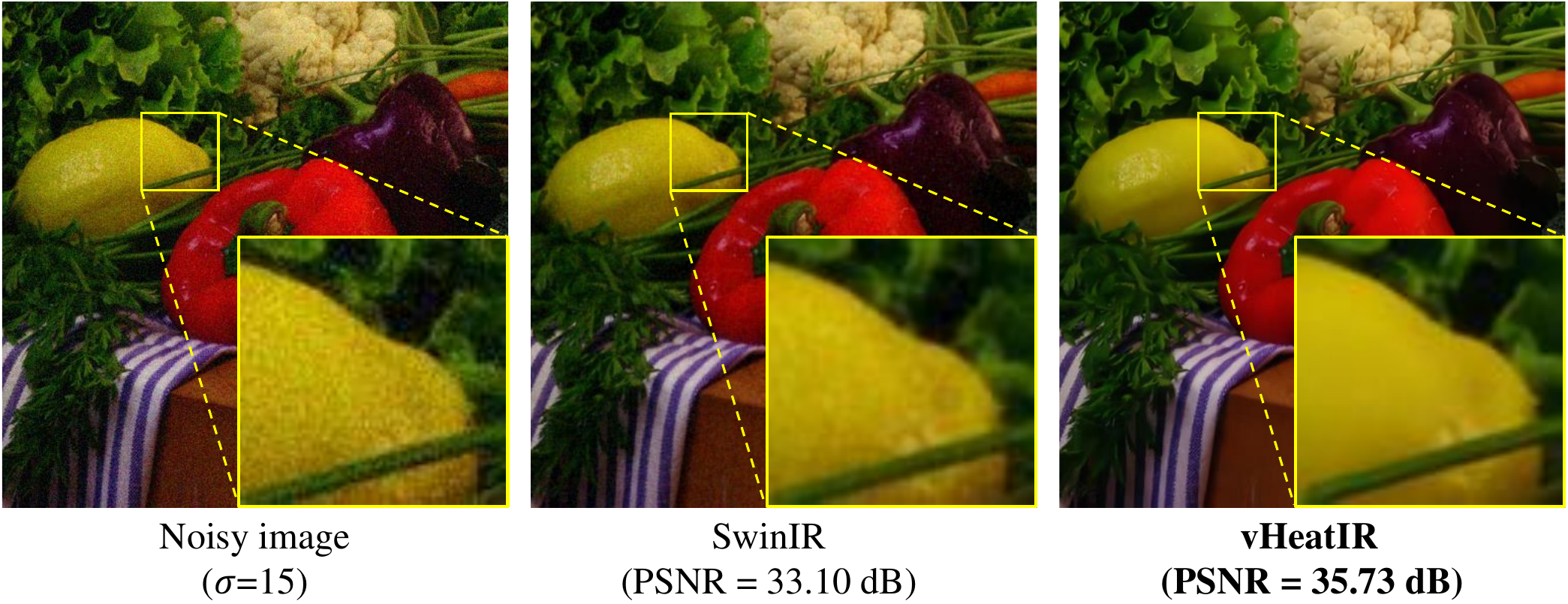}
\caption{Color image denoising visualization of vHeatIR and SwinIR after 15000 training iterations ($\sigma=15$). The input image is selected from McMaster~\citep{mcmaster}.}
\label{fig:lowlevel}
\end{center}
\vspace{-15pt}
\end{figure}

\begin{table}
\setlength{\tabcolsep}{0.15cm}
\centering
    \caption{Quantitative comparison (average PSNR) on low-level vision tasks. $^\dagger$: results are reproduced for a fair comparison.}
    %\vspace{-15pt}
    % \scalebox{0.78}{
    \begin{tabular}{lccc}
        \toprule
        \multirow{3}{*}{Model} & Image Denoising & JPEG Compression \\
        &(Set12/McMaster, & Artifact Reduction  \\
        &$\sigma=15$)  & (LIVE1, $q=40$) \\
        \midrule
        DnCNN~\citep{dncnn} & 32.86/33.45 & 33.96 \\
        DRUNet~\citep{drunet} & 33.25/35.40 & 34.58 \\
        SwinIR$^\dagger$~\citep{swinir} & 33.33/35.55 & 34.61 \\
        %\rowcolor{Gray}
        \rblue
        vHeatIR (Ours) & \textbf{33.37/35.60} & \textbf{34.64} \\
        %ViT-B & 26.2 & 35.7 & 49.3 & 36.4 & 49.7 \\
        \bottomrule
    \end{tabular}
    % }
    \label{tab: low-level}
\vspace{-5pt}
\end{table}

\noindent\textbf{Low-level vision tasks.} To further evaluate the generalization capability of our proposed vHeat model, we integrate the Heat Conduction Operator (HCO) by replacing the self-attention modules in the SwinIR~\citep{swinir} model, resulting in the vHeatIR architecture. We then conduct a series of experiments on several standard low-level vision tasks to assess the performance of vHeatIR. These tasks include grayscale and color image denoising on the Set12~\citep{set12} and McMaster~\citep{mcmaster} datasets, as well as JPEG compression artifact reduction on the LIVE1~\citep{live1} dataset. In these experiments, we use the same settings as those in SwinIR to ensure a fair comparison. The results, as summarized in Table~\ref{tab: low-level}, demonstrate that vHeatIR consistently outperforms the other baseline models. This improvement is largely attributed to the ability of HCO to operate efficiently in the frequency domain, which enhances the model's performance in handling low-level image details. After training for 15,000 iterations, we visualize the denoising results on color images with noise level \(\sigma = 15\) in Fig.~\ref{fig:lowlevel}. As shown, vHeatIR produces noticeably cleaner images compared to SwinIR, indicating its superior ability to restore image quality. These results not only highlight the effectiveness of the proposed vHeat model but also validate its strong generalization capabilities for low-level vision tasks.

\subsection{Analysis of Dynamic Locality}

\begin{figure}[t]
\centering
\begin{center}
\includegraphics[width=.47\textwidth]{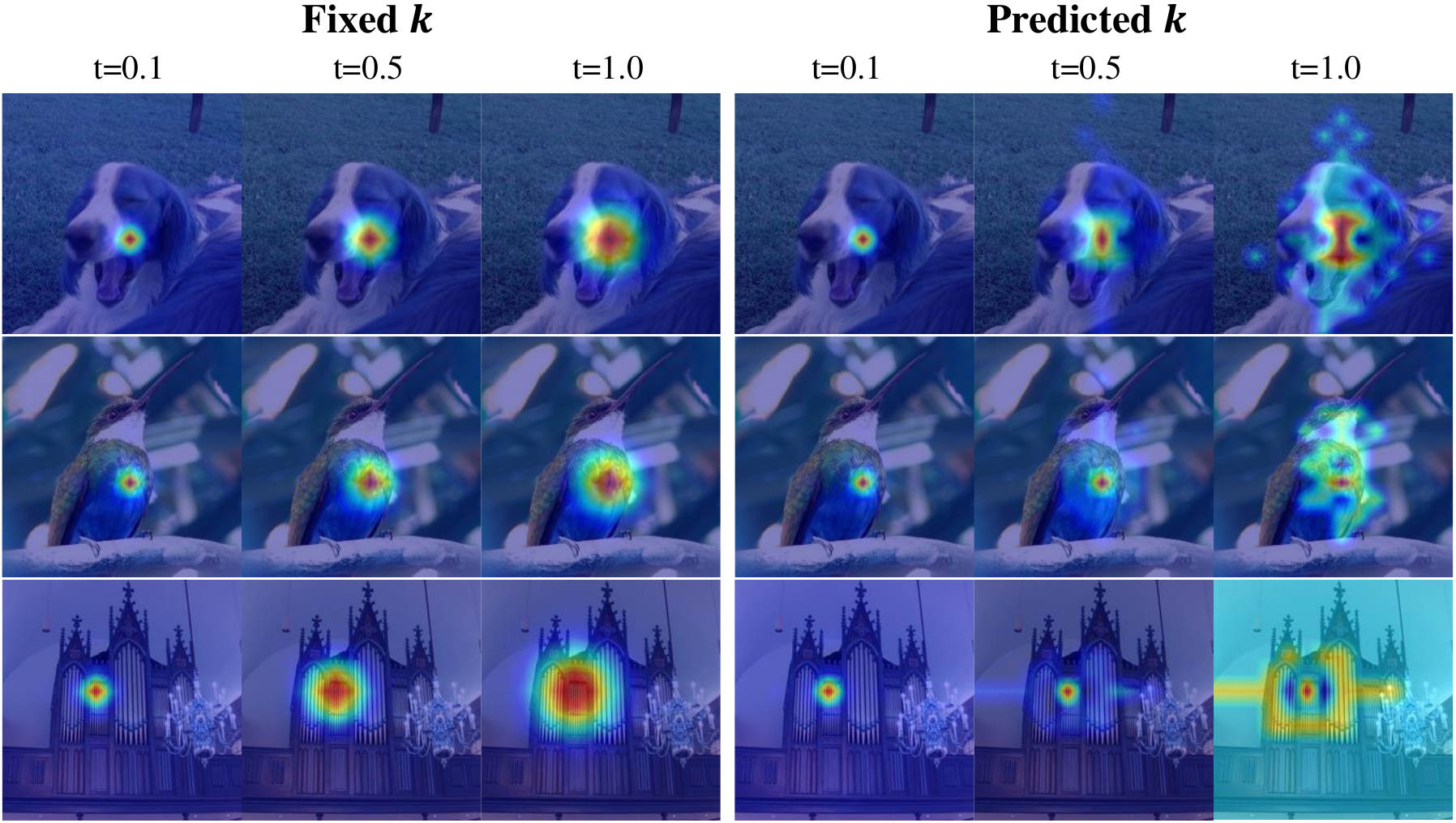}
%\vspace{-10pt}
\caption{Temperature distribution ($U^t$) when using a randomly selected patch as the heat source. \textit{Please zoom in for details.}
%
%vHeat-B model at time $t=0.1/0.5/1.0$ when the thermal diffusivity $k$ is fixed (Left) or predicted (Right). Predicted $k$ brings self-adaptive visual heat conduction to vHeat.
} %选取图像中的一个patch作为热源，可视化了在fixed k和predicted k情况下，stage2某层，t=0.1/0.5/1.0时刻的U^t，表示了该patch对于全图的响应
\label{fig:vis_vheat}
\end{center}
\vspace{-15pt}
\end{figure}

\noindent\textbf{Visual Heat Conduction.}
%
%To validate the nonuniform visual heat conduction brought by $k$,
%in the pre-trained vHeat-
The proposed vHeat works upon an adaptive filtering mechanism. To verify this claim, in Fig.~\ref{fig:vis_vheat}, we visualize the temperature $U^t$ defined in ~\eqref{eq:implementation_U} under predicted $k$ when a random patch is taken as the heat source. 
%
%This shows the correlation of the selected patch to the whole image.
%This shows the correlation of the patch to the whole image of the selected patch.
%
%$k$ delivers the standard isotropic heat conduction, 
With a predicted $k$, vHeat delivers self-adaptive visual heat conduction.
As the heat conduction time ($t$) increases, the correlation between the selected patch and the entire image improves, which effectively filters out unrelated patches in the frequency domain. Please refer to Sec. \ref{appendix:erf_vis} in the supplementary for vHeat's effective receptive field visualization. 

%, which indicates predicting $k$ by FVEs has provided the adaptability of visual representation to vHeat.

%{\color{red}
\noindent\textbf{Ablation of thermal diffusivity.} To demonstrate the effectiveness of shared FVEs, we conduct the following experiments on ImageNet-1K: (1) Fix the thermal diffusivity \(k = 0.0/1.0/10.0\), (2) Treat \(k\) as a learnable parameter for each layer, and (3) Use individual FVEs to predict \(k\) for each layer. As shown in Table~\ref{table:k}, when \(k = 0.0\), the visual heat conduction does not function effectively. A larger fixed \(k\) value, e.g., \(k = 5.0\), allows HCO to operate isotropically without considering the image content, achieving a top-1 accuracy of 81.7\%. Predicting \(k\) via FVEs outperforms treating \(k\) as a learnable parameter, likely due to the enhanced prior knowledge of frequency values provided by FVEs. After performing DCT, the features lose explicit frequency values, whereas FVEs offer the model prior frequency information. Similar to how positional encoding enhances performance in models that already incorporate positional information~\citep{cnnseq2seq}, predicting \(k\) using FVEs rather than treating it as a learnable parameter reinforces the frequency prior and clarifies the relationship between frequency and thermal diffusivity. When \(k\) is predicted by shared FVEs, the performance improves to 82.2\%, validating that shared FVEs effectively reduce learning diffusivity and further enhance performance.
%}

\begin{table}[t]
\setlength{\tabcolsep}{0.13cm}
\caption{Evaluating thermal diffusivity $k$ with vHeat-T.}
\vspace{-4pt}
\centering
% \scalebox{0.85}{
    \begin{tabular}{lc}
        \toprule
        Settings & top-1 acc. (\%) \\ 
        \midrule
        Fixed $k=0.0$ & 81.0 \\
        Fixed $k=1.0$ & 81.7 \\
        Fixed $k=5.0$ & 81.8 \\        
        $k$ as a learnable parameter & 81.5 \\
        Predicting $k$ using individual FVEs & 82.0 \\
        Predicting $k$ using shared FVEs & \textbf{82.2} \\
        \bottomrule
    \end{tabular}
    % }
\label{table:k}
%\vspace{-2.5pt}
\end{table}

% \begin{table*}[h]
% \vspace{-10pt}
% \caption{\textbf{Left}: Evaluating thermal diffusivity $k$ with vHeat-T. \textbf{Right}: Comparison of vHeat with global filters, where vHeat-B$^\star$ denotes replacing HCOs in vHeat-B with operators proposed in GFNet.}
%     \resizebox{0.515\textwidth}{!}{
%     \begin{tabular}{lc}
%         \toprule
%         Settings & top-1 acc. (\%) \\ 
%         \midrule
%         Fixed $k=0.0$ & 81.0 \\
%         Fixed $k=1.0$ & 81.7 \\
%         Fixed $k=5.0$ & 81.8 \\        
%         $k$ as a learnable parameter & 81.5 \\
%         Predicting $k$ using individual FVEs & 82.0 \\
%         Predicting $k$ using shared FVEs & \textbf{82.2} \\
%         \bottomrule
%     \end{tabular}
%     }
%     \resizebox{0.475\textwidth}{!}{
%     \begin{tabular}{lcc|c}
%         \toprule
%         Model & \#Param. & FLOPs & top-1 acc. (\%) \\
%         \midrule
%         GFNet-H-B & 54M & 8.4G & 82.9 \\
%         vHeat-S & 50M & 8.5G & \textbf{83.6} \\
%         %vHeat-B & 87M & 14.9G & 83.6 \\
%         \midrule
%         vHeat-B$^\star$ & 68M & 11.2G & 83.5 \\
%         vHeat-B & 68M & 11.2G & \textbf{84.0} \\
%         %Plain vHeat-B & 88M & & 82.4 \\
%          %& 88M & 16.9G & \textbf{82.6} \\
%         \bottomrule
%     \end{tabular}
%     }
%     \label{tab: ablation}
%     \vspace{-10pt}
% \end{table*}

\begin{table}
\setlength{\tabcolsep}{0.3cm}
\centering
\caption{Comparison of vHeat with global filters, where vHeat-B$^\star$ denotes replacing HCOs in vHeat-B with GFNet operators.}
\vspace{-6pt}
\centering
% \scalebox{0.87}{
    \begin{tabular}{lcc|c}
        \toprule
        Model & \#Param. & FLOPs & top-1 acc. (\%) \\
        \midrule
        GFNet-H-B & 54M & 8.4G & 82.9 \\
        vHeat-S & 50M & 8.5G & \textbf{83.6} \\
        %vHeat-B & 87M & 14.9G & 83.6 \\
        \midrule
        vHeat-B$^\star$ & 68M & 11.2G & 83.5 \\
        vHeat-B & 68M & 11.2G & \textbf{84.0} \\
        %Plain vHeat-B & 88M & & 82.4 \\
         %& 88M & 16.9G & \textbf{82.6} \\
        \bottomrule
    \end{tabular}
    % }
\label{table:gfnet}
\vspace{-5pt}
\end{table}

\subsection{Comparison With Global Filters}

To simulate physical heat conduction in vision tasks, we developed the Heat Conduction Operator (HCO) in the frequency domain. We compare HCO with (1) GFNet~\citep{gfnet}, a model using global filters in the frequency domain, and (2) an ablation where HCOs are replaced by GFNet's frequency-domain operators. The results in Table~\ref{table:gfnet} show that vHeat-S outperforms GFNet-H-B with a similar model size. Furthermore, when HCOs are substituted with GFNet's operators, there is a noticeable performance drop, underscoring the unique advantages of the proposed HCO. These findings validate the efficacy of our proposed HCO and the underlying visual heat conduction modeling in enhancing feature representation.

%To systematically simulate physical heat conduction in vision tasks, we developed the Heat Conduction Operator (HCO), which operates in the frequency domain. Given its design, we conduct a comparative study to evaluate HCO's effectiveness. Specifically, we compare HCO with (1) GFNet~\citep{gfnet}, a vision representation model that leverages global filters in the frequency domain, and (2) an ablation setting where HCO is replaced with the frequency-domain operators proposed in GFNet. The results, summarized in Table~\ref{table:gfnet}, demonstrate that vHeat-S outperforms GFNet-H-B under a similar model size, showcasing superior performance. Furthermore, when HCO is substituted with GFNet's operators, there is a noticeable performance drop, underscoring the unique advantages of the proposed HCO. These findings validate the efficacy of HCO and the underlying visual heat conduction modeling in enhancing feature representation and improving model performance across various vision tasks.

\section{Conclusion}
\label{sec:conclusion}
%We introduce vHeat, a visual representation model that combines the benefits of global receptive fields, computational efficiency, and enhanced interpretability. The effectiveness of the vHeat model family, including vHeat-T/S/B models, has been demonstrated through extensive experiments and ablation studies, significantly outperforming popular CNNs and ViTs. The results highlight the potential of vHeat as a new paradigm for vision representation learning, offering fresh insights for the development of physics-inspired visual representation models.
%
We introduce vHeat, a visual representation model that integrates the advantages of global receptive fields, computational efficiency, and enhanced interpretability. Extensive experiments and ablation studies demonstrate the efficiency and effectiveness of the vHeat model family, including vHeat-T/S/B models, which significantly outperform popular CNNs and ViTs. These results highlight vHeat's potential as a novel paradigm for vision representation learning, offering fresh insights for the development of physics-inspired representation models in computer vision.

\newpage

\section{Acknowledgement}

This work was supported by National Natural Science Foundation of China (NSFC) under Grant 62225208 and 62450046 and CAS Project for CAS Project for Young Scientists in Basic Research under Grant No.YSBR-117. 

{
    \small
    \bibliographystyle{ieeenat_fullname}
    \bibliography{main}
}
%\appendix
\clearpage
\onecolumn

\appendix

\section{Motivation}

Modern visual representation models are built upon the attention mechanism inspired by biological vision systems. One drawback of it is the lack of a clear definition of the relationship between biological electrical signals and brain activity (energy). This drives us to break through the attention mechanism and attempt other physical laws. Heat conduction is a physical phenomenon in nature, characterized by the propagation of energy. The heat conduction process combines implicit attention computation with energy computation and has the potential to be a new mechanism for visual representation models.

\section{HCO implementation using $\text{DCT}_\text{2D}$ and $\text{IDCT}_\text{2D}$}
\label{appendix: DCT}

Assume a matrix denoted as $\mathbf{A}$ and the transformed matrix denoted as $\mathbf{B}$, the $\mathbf{DCT_{2D}}$ and the $\mathbf{IDCT_{2D}}$ can be performed by
\begin{equation}
\label{eq:DCT}
\begin{aligned}
&\mathbf{DCT_{2D}}: \mathbf{B}_{pq}=\mathbf{\alpha_p} \mathbf{\alpha_q}\sum_{m=0}^{M-1}\sum_{n=0}^{N-1} \mathbf{A}_{mn} \mathbf{cos}\frac{(2m+1)p\pi}{2M} \mathbf{cos}\frac{(2n+1)q\pi}{2N},\\
&\mathbf{IDCT_{2D}}: \mathbf{A}_{mn}=\sum_{m=0}^{M-1}\sum_{n=0}^{N-1} \mathbf{\alpha_p} \mathbf{\alpha_q} \mathbf{B}_{pq} \mathbf{cos}\frac{(2m+1)p\pi}{2M} \mathbf{cos}\frac{(2n+1)q\pi}{2N},
\end{aligned}
\end{equation}
where $0{\leq}\{p,m\}{\leq}M-1$, $0{\leq}\{q,n\}{\leq}N-1$, $\mathbf{\alpha_p}=\left\{\begin{aligned}&\frac{1}{\sqrt{M}},p=0\\&\frac{2}{\sqrt{M}},p>0\end{aligned}\right.$, and $\mathbf{\alpha_q}=\left\{\begin{aligned}&\frac{1}{\sqrt{N}},q=0\\&\frac{2}{\sqrt{N}},q>0\end{aligned}\right.$.
$M$ and $N$ respectively denote the row and column sizes of $\mathbf{A}$.
Considering the matrix multiplication is GPU-friendly, we implement the $\mathbf{DCT_{2D}}$ and $\mathbf{IDCT_{2D}}$ in Eq. ~\eqref{eq:DCT} by
\begin{equation}
\label{eq:matrix_DCT}
\begin{aligned}
\mathbf{C}&=(\mathbf{C}_{mp})_{M\times{M}}=\left(\mathbf{\alpha_{p}} \mathbf{cos}\frac{(2m+1)p\pi}{2M}\right)_{M\times{M}},\\
\mathbf{D}&=(\mathbf{D}_{nq})_{N\times{N}}=\left(\mathbf{\alpha_{q}} \mathbf{cos}\frac{(2n+1)q\pi}{2N}\right)_{N\times{N}},\\
\mathbf{B}&=\mathbf{CAD^T},\\
\mathbf{A}&=\mathbf{C^TBD}.
\end{aligned}
\end{equation}
% Considering the fast speed of matrix multiplication on GPUs, we utilize Eq.\eqref{eq:matrix_DCT} for implementation. 
Suppose the number of total patches is $N$ and the image is square, the shapes of $\mathbf{A}$, $\mathbf{B}$, $\mathbf{C}$ and $\mathbf{D}$ are all $\sqrt{N}\times\sqrt{N}$, which illustrates the computational complexity of \eqref{eq:matrix_DCT} and HCO is $O(N^{1.5})$. 

We compared our implementation of DCT/IDCT in vHeat with Torch-DCT, which is implemented based on $torch.fft$. Our implemented vHeat-B (661 img/s) is much faster than Torch-DCT (367 img/s), validating that our implemented GPU-friendly matrix multiplication is significantly efficient.

\section{Experimental Settings}
\label{appendix: settings}

\textbf{Model configurations.} The configurations of vHeat-T/S/B models are shown in Table ~\ref{tab: model_cfg}. The FLOPs and training parameters are reported after reparameterization in HCOs.

\begin{table*}[h]
    \begin{center}
    \caption{Configurations of vHeat. The contents in the tuples represent configurations for four stages.}
    \label{tab: model_cfg}
    \scalebox{1.0}{
    \begin{tabular}{l|c|c|c}
        \toprule
        %\multicolumn{3}{c}{\textbf{Mask R-CNN 1$\times$ schedule on COCO}}\\
        %\midrule
        Size & Tiny & Small & Base \\ 
        \midrule
        Stem & \multicolumn{3}{c}{3$\times$3 conv with stride 2; Norm; GELU; 3$\times$3 conv with stride 2; Norm} \\
        Downsampling & \multicolumn{3}{c}{3$\times$3 conv with stride 2; Norm} \\
        MLP ratio & \multicolumn{3}{c}{4} \\
        Classifier head & \multicolumn{3}{c}{Global average pooling, Norm, MLP
} \\\midrule
        Layers & \begin{tabular}[c]{@{}c@{}} (2,2,6,2) (classification) \\ (2,2,5,2) (others)\end{tabular} & \begin{tabular}[c]{@{}c@{}} (2,2,18,2) (classification) \\ (2,2,16,2) (others)\end{tabular} & \begin{tabular}[c]{@{}c@{}} (2,2,18,2) (segmentation) \\ (4,4,20,4) (others)\end{tabular} \\  
        \midrule
        Channels & (96,192,384,768) & (96,192,384,768) & \begin{tabular}[c]{@{}c@{}} (128,256,512,1024) (segmentation) \\ (96,192,384,768) (others)\end{tabular} \\
        \bottomrule
    \end{tabular}
    }
    \end{center}
\end{table*}

\textbf{Image Classification.}
\label{sec:imagenet1k}
%
%We first assess vHeat's capability of image classification on the ImageNet-1K dataset~\citep{ImageNet2009}. 
Following the standard evaluation protocol used in~\citep{liu2022swin}, all vHeat series are trained from scratch for $300$ epochs and warmed up for the first $20$ epochs. We utilize the AdamW optimizer~\citep{adamw} during the training process with betas set to $(0.9, 0.999)$, a momentum of $0.9$, a cosine decay learning rate scheduler, an initial learning rate of $2\times10^{-3}$, a weight decay of $0.08$, and a batch size of $2048$. The drop path rates are set to $0.1/0.3/0.5$ for vHeat-T/S/B, respectively.  Other techniques such as label smoothing ($0.1$) and exponential moving average (EMA) are also applied. No further training techniques are employed beyond these for a fair comparison. The training of vHeat-T/S/B takes 4.5/7/8.5 minutes per epoch on Tesla 16$\times$V100 GPUs. 

\textbf{Object Detection.}
Following the settings in Swin~\citep{liu2022swin} with the Mask-RCNN detector, we build the vHeat-based detector using the MMDetection library~\citep{MMdet2019}. 
The AdamW optimizer~\citep{adamw} with a batch size of $16$ is used to train the detector. 
The initial learning rate is set to $1\times 10^{-4}$ and is reduced by a factor of $10\times$ at the 9th and 11th epoch.
The fine-tune process takes $12$ ($1\times$) or $36$ ($3\times$) epochs. 
We employ the multi-scale training and random flip technique, which aligns with the established practices for object detection evaluations.
%
%0510，需要补充加载vheat参数时的插值方式，及相关的对比实验，放在ablation or appendix

\textbf{Semantic Segmentation.} 
Following the setting of Swin Transfomer~\citep{Swin2021}, we construct a UperHead~\citep{upernet} on top of the pre-trained vHeat model to test its capability for semantic segmentation. 
The AdamW optimizer~\citep{adamw} is employed and the learning rate is set to $6\times10^{-5}$ with a batch size of $16$. 
The fine-tuning process takes a total of standard $160k$ iterations and the default input resolution is $512\times512$. 

\section{Additional Ablation Studies}

\begin{table}[!h]
    \begin{center}
    \caption{Evaluating different methods to align the shape of FVEs/$k$ when loading ImageNet-1K pre-trained vHeat-B weights for detection and segmentation on COCO.
    }
    \label{tab: interpolate_ablation}
    \begin{tabular}{lcc}
        \toprule
        %\multicolumn{3}{c}{\textbf{Mask R-CNN 1$\times$ schedule on COCO}}\\
        %\midrule
        Method & AP$^\text{b}$ & AP$^\text{m}$ \\ 
        \midrule
        Interpolating FVEs to predict $k$ & 47.4 & 42.9 \\
        Adding 0 to FVEs  & 47.4 & 42.7 \\
        Adding 0, then interpolating FVEs & \textbf{47.7} & \textbf{43.0} \\        
        Interpolating the predicted $k$ & 47.2 & 42.7 \\
        \bottomrule
    \end{tabular}
    \end{center}
\end{table}

\subsection{Interpolation of FVEs/$k$ for downstream tasks}
\label{appendix: interpolation}

We have tried several approaches to align the shape for ablation. (1) Directly interpolate FVEs to the target shape of the input image. (2) Add 0 to the lower right region of FVEs to align the target shape. (3) Add 0 to the lower right region of FVEs to $512\times512$, and interpolate to the target shape. (4) Directly interpolate the predicted thermal diffusivity $k$ to the target shape. The results are summarized in Table ~\ref{tab: interpolate_ablation}. Through the comparison, we select adding 0, then interpolating FVEs to the target shape for all downstream tasks.

\subsection{Plain vHeat model}
\label{appendix: plain}

We've tested the performance of plain vHeat-B on ImageNet-1K classification. Keeping the same as DeiT-B, plain vHeat-B has 12 HCO layers, 768 embedding channels and the patch size is set to 16. Results are shown in Table ~\ref{tab: plain}. The superiority of plain vHeat-B over DeiT-B also validates the effectiveness of vHeat model. 

\begin{table}[!h]
    \begin{center}
    \caption{Plain vHeat-B vs. DeiT-B on ImageNet-1K with 300 epochs supervised training. %Evaluating different methods to align the shape of FVEs when loading ImageNet-1K pre-trained vHeat-B weights for detection and segmentation on COCO.
    }
    \label{tab: plain}
    \begin{tabular}{lcc|c}
        \toprule
        %\multicolumn{3}{c}{\textbf{Mask R-CNN 1$\times$ schedule on COCO}}\\
        %\midrule
        Model & \#Param. & FLOPs & Acc \\ 
        \midrule
        DeiT-B & 86M & 17.5G & 81.8 \\
        %Plain vHeat-B & 88M & & 82.4 \\
        Plain vHeat-B & 88M & 16.9G & \textbf{82.6} \\
        \bottomrule
    \end{tabular}
    \end{center}
\end{table}

\subsection{Depth-wise convolution}
\label{appendix: dwconv}

We conduct experiments to validate the performance improvement from DWConv. We replace depth-wise convolution with layer normalization for vHeat-B. Results are summarized in Table ~\ref{tab: dw}, and vHeat-B achieves 83.8\% Top-1 accuracy on ImageNet-1K classification, 0.2\% lower than with DWConv, which validates the main gains come from the proposed HCO. Besides, when $k$ is fixed as a large value, e.g. $k=10.0$, replacing DWConv with layer normalization causes a significant performance drop (-0.7\% top-1 accuracy). The comparison validates predicting $k$ by FVEs can effectively improve the robustness of vHeat. 

\begin{table}[!h]
    \begin{center}
    \caption{Ablation experiments of depth-wise convolution (DWConv).}
    \label{tab: dw}
    \begin{tabular}{lc|c}
        \toprule
        Model & DWConv & Acc \\
        \midrule
        vHeat-B & \checkmark & 84.0 \\
        vHeat-B & \ding{55}  & 83.8 (-0.2) \\
        %vHeat-B & 87M & 14.9G & 83.6 \\
        \midrule
        vHeat-B (fix $k$=10.0) & \checkmark & 83.6 \\
        vHeat-B (fix $k$=10.0) & \ding{55} & 82.9 (-0.7) \\
        \bottomrule
    \end{tabular}
    \end{center}
\end{table}

% \subsection{Predicting $k$ by FVEs \textit{vs.} treating $k$ as a learnable parameter}
% \label{appendix: k_detail_analysis}

% After performing DCT, the features lack explicit frequency value, while FVEs provide the model with prior knowledge of frequency values. Similar to how the introduction of positional encoding can enhance performance even in models that include positional information~\citep{cnnseq2seq}, predicting $k$ by FVEs, rather than treating $k$ as a learnable parameter, reinforces prior frequency information and more clearly represents the relationship between frequency and thermal diffusivity. 

\section{Receptive Field Visualization}
\label{appendix:erf_vis}
\begin{figure}[h]
\centering
\begin{center}
\includegraphics[width=0.8\textwidth]{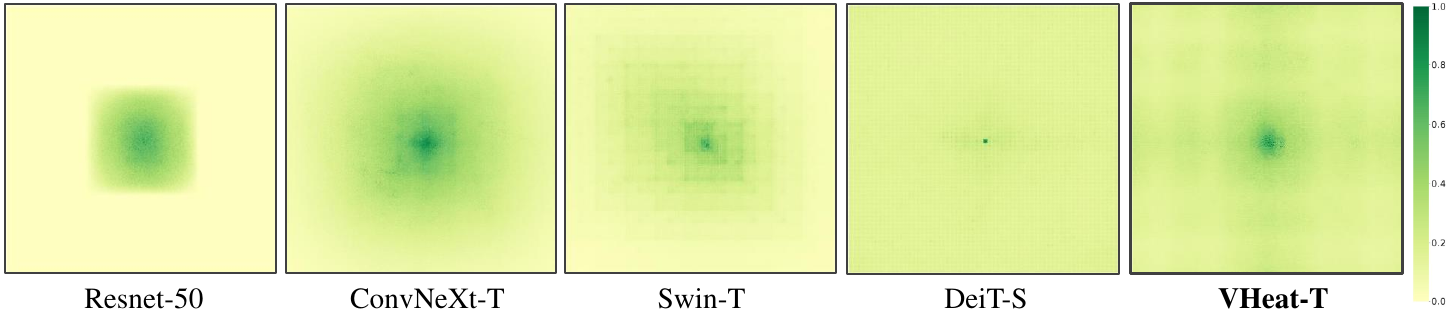}
%\vspace{-10pt}
\caption{Visualization of the effective receptive fields (ERF)~\citep{luo2016understanding}.
%for ResNet50~\citep{Resnet2016}, ConvNeXt-T~\citep{liu2022convnet}, Swin-T~\citep{Swin2021}, DeiT-S~\citep{DeiT2021} (ViT), HiViT-T~\citep{hivit}, and vHeat-T. 
The visualization of baseline models are provided from VMamba~\citep{liu2024vmamba}. Pixels of higher intensity indicate larger responses with the central pixel.}
\label{fig:erf}
\end{center}
\end{figure}

The Effective Receptive Field (ERF)~\citep{luo2016understanding} of an output unit denotes the region of input that contains elements with a non-negligible influence on that unit. 
In Fig.~\ref{fig:erf}, ResNet, ConNeXT, and Swin have local ERF. DeiT~\citep{DeiT2021} and vHeat exhibit global ERFs. The difference lies in that DeiT has a $\mathcal{O}(N^2)$ complexity while vHeat enjoys $\mathcal{O}(N^{1.5})$ complexity.

\section{Heat Conduction Visualization}
\label{more_vis}

\begin{figure}[h]
    \centering
    \includegraphics[width=0.6\textwidth]{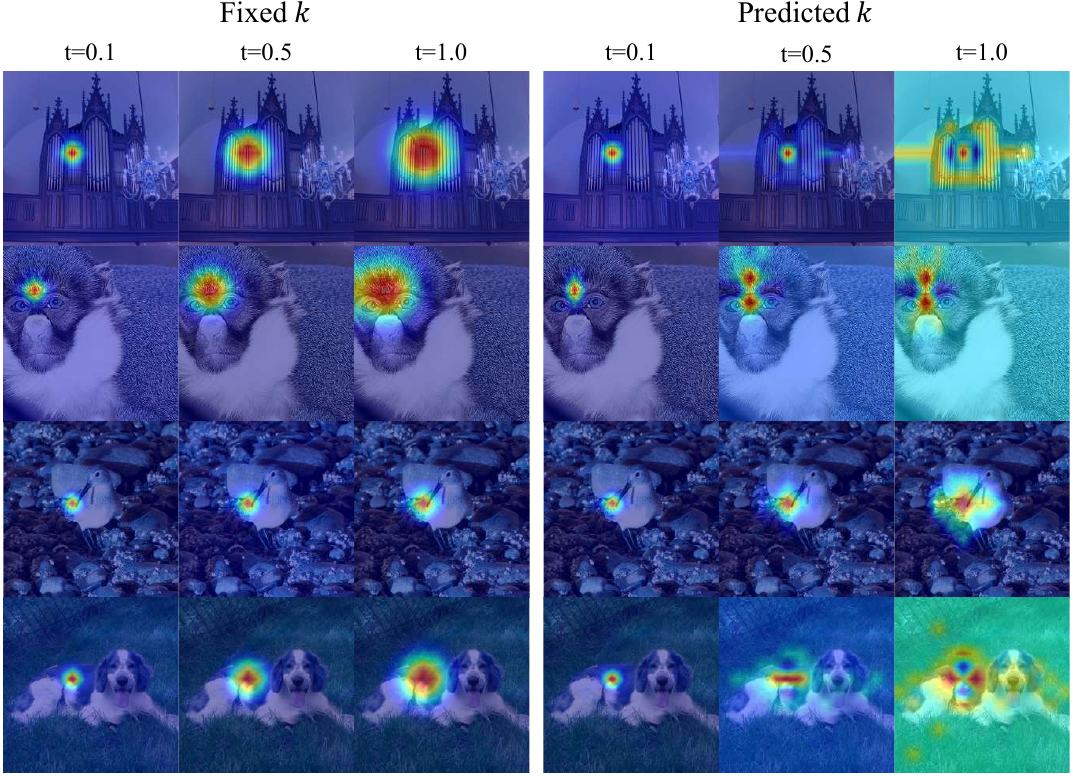}
    \caption{Temperature distribution ($U^t$) when using a randomly selected patch as the heat source. (Best viewed in color)}
    \label{fig:more_vis}
\end{figure}

We visualize more instances of visual heat conduction, given a randomly selected patch as the heat source, Fig. ~\ref{fig:more_vis}, validating the self-adaptive visual heat conduction pattern through the prediction of $k$. 

\section{Analysis of $k$ in each layer}

\begin{figure}[h]
    \centering
    \includegraphics[width=0.5\textwidth]{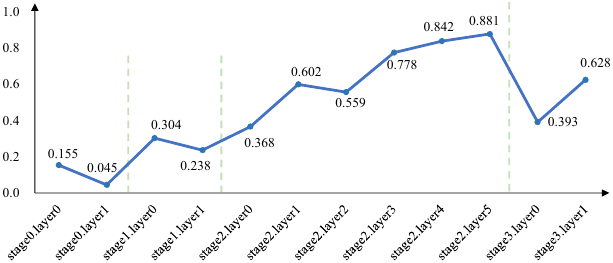}
    \caption{Average values of $k$ in each layer.}
    \label{fig:kmean}
\end{figure}

We calculate average values of $k$ in each layer of ImageNet-1K classification pre-trained vHeat-Tiny, Fig. ~\ref{fig:kmean}. In stage 2 and stage 3, average values of $k$ corresponding to deeper layers are larger, indicating that the visual heat conduction effect of deeper layers is stronger, leading to faster and farther overall content propagation. 

\section{Feature Map Visualization}

\begin{figure}[h]
    \centering
    \includegraphics[width=0.4\textwidth]{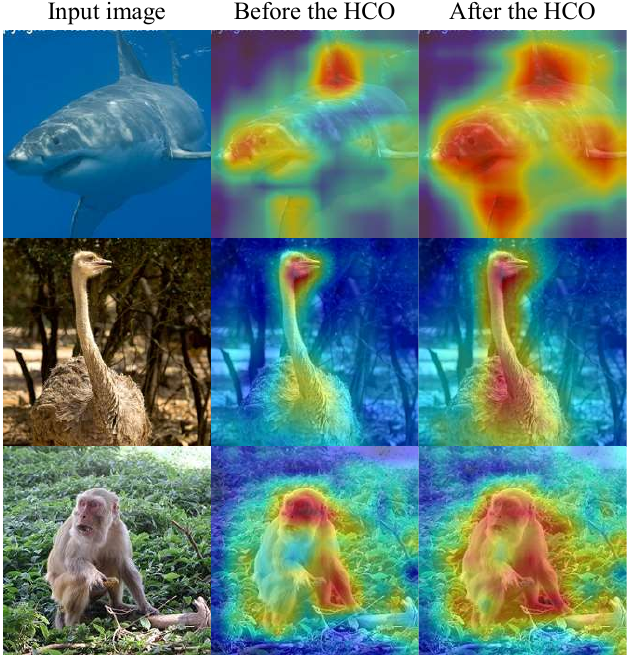}
    \caption{Visualization of the feature before/after HCO in a random layer in stage 2 with ImageNet-1K classification pre-trained vHeat-B. The images are randomly selected from ImageNet-1K. }
    \label{fig:feature}
\end{figure}

We visualize the feature before/after HCO in a random layer in stage 2 with randomly selected images as input, Fig. ~\ref{fig:feature}. Before HCO, only a few regions of the foreground object are activated. After HCO, almost the entire foreground object is activated intensively.

\section{Ablation of the Linear+SiLU branch}

The Linear + SiLU branch is a gated MLP unit, which is inspired by the design used in LLaMA~\citep{llama}. We conducted ablation experiments, and the results, as shown in Table~\ref{tab:r1a2}, demonstrate that the primary performance improvement of vHeat comes from the HCO, rather than the gated MLP.

\begin{table}[h]
    \centering
    \begin{tabular}{ccccc}
        \toprule
        Model & \#Param. & FLOPs & Top-1 acc. (\%) \\
        \midrule
         vHeat-B & 68M & 11.2G & 84.0 \\
         Linear + ReLU & 68M & 11.2G & 83.9 \\
         w/o (Linear + SiLU) & 62M & 10.3G & 83.6 \\
         w/o HCO & 49M & 8.0G & 76.7 \\
         \bottomrule
    \end{tabular}
    \caption{Ablation study of the Linear + SiLU branch.}
    \label{tab:r1a2}
\end{table}

\section{Comparison with SOTAs}

\begin{figure}
    \centering
    \includegraphics[width=0.7\linewidth]{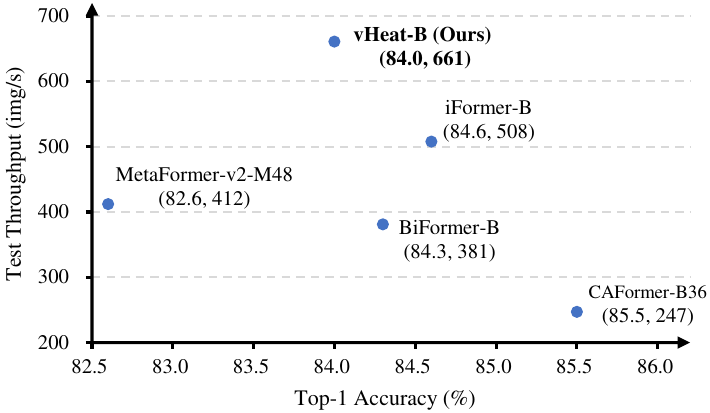}
    \caption{Comparison of vHeat-B and other base-level SOTA vision models.}
    \label{fig:supp_comparison}
\end{figure}

We compare vHeat-B with other base-level SOTA visual representation models (MetaFormer-v2-M48, CAFormer-B36~\citep{metaformer}, iFormer-B~\citep{iformer}, and BiFormer-B~\citep{biformer}) on Top-1 Accuracy on ImageNet-1K and test throughput (an A100 GPU with 128 batch size), Figure~\ref{fig:supp_comparison}. Our proposed vHeat demonstrates comparable performance with substantially improved test throughput.

\end{document}